\documentclass[journal]{IEEEtran}
\ifCLASSINFOpdf
\else
\fi
\usepackage[numbers,sort&compress]{natbib}
\usepackage{color}
\usepackage{amsmath}
\usepackage{amsfonts,amssymb}
\usepackage{multirow}
\usepackage{graphicx}
\usepackage{epstopdf}
\usepackage{float}
\usepackage{subfigure}

\begin{document}
%
\title{HAAP: Vision-context Hierarchical Attention Autoregressive with Adaptive Permutation for Scene Text Recognition}
%
%
%

\author{Honghui Chen,
        Yuhang Qiu,
        Jiabao Wang,
        Pingping Chen,~\IEEEmembership{Senior~Member,~IEEE,}
        and~Nam Ling,~\IEEEmembership{Life~Fellow,~IEEE}
\thanks{This work was supported by the NSFC (No. 62171135), Fujian Key Project (No.2023XQ004), Fujian  Eagle-Scholar and Distinguished Project (No. 2022J06010), Fujian Natural Science Fund (No. 2024H6013).(Corresponding author: Pingping Chen.)}
\thanks{Honghui Chen, Jiabao Wang, Pingping Chen are with the College of Physics and Information Engineering, Fuzhou University, Fuzhou, 350108, China (e-mail: chh5840996@gmail.com; wabbb0811@163.com; ppchen.xm@gmail.com).}
\thanks{Yuhang Qiu is with the Faculty of Engineering, Monash University, Clayton, VIC, 3800, Australia (e-mail: yuhang.qiu@monash.edu).}
\thanks{Nam Ling is with the Department of Computer Science and Engineering, Santa Clara University, Santa Clara, California, 95053, USA (e-mail: nling@scu.edu).}
\thanks{Manuscript received January 31, 2024; revised May 22, 2025.}}

%
%

\markboth{Journal of \LaTeX\ Class Files,~Vol.~14, No.~8, January~2024}%
{Shell \MakeLowercase{\textit{et al.}}: Bare Demo of IEEEtran.cls for IEEE Journals}
%



\maketitle

\begin{abstract}
Scene Text Recognition (STR) is challenging in extracting effective character representations from visual data when text is unreadable. Permutation language modeling (PLM) is introduced to refine character predictions by jointly capturing contextual and visual information. However, in PLM, the use of random permutations causes training fit oscillation, and the iterative refinement (IR) operation also introduces additional overhead. To address these issues, this paper proposes the Hierarchical Attention autoregressive Model with Adaptive Permutation (HAAP) to enhance position-context-image interaction capability, improving autoregressive LM generalization. First, we propose Implicit Permutation Neurons (IPN) to generate adaptive attention masks that dynamically exploit token dependencies, enhancing the correlation between visual information and context. Adaptive correlation representation helps the model avoid training fit oscillation. Second, the Cross-modal Hierarchical Attention mechanism (CHA) is introduced to capture the dependencies among position queries, contextual semantics and visual information. CHA enables position tokens to aggregate global semantic information, avoiding the need for IR. Extensive experimental results show that the proposed HAAP achieves state-of-the-art (SOTA) performance in terms of accuracy, complexity, and latency on several datasets.
\end{abstract}


%
\IEEEpeerreviewmaketitle

\section{Introduction}
%
%
%
%
\IEEEPARstart{S}{cene} Text Recognition (STR) aims to transcribe text in an image to a computer-readable text format, i.e., to recognize the localized text regions. Scene text with rich information plays a vital role in a range of applications such as visual quizzing, autonomous driving, image retrieval, augmented reality, retail, education, and visually impaired devices \cite{song2022clip, taki2023scene, luo2022clip4clip, 9798797, 10078345}.

\begin{figure}[]
\centering
\includegraphics[width=8.5 cm]{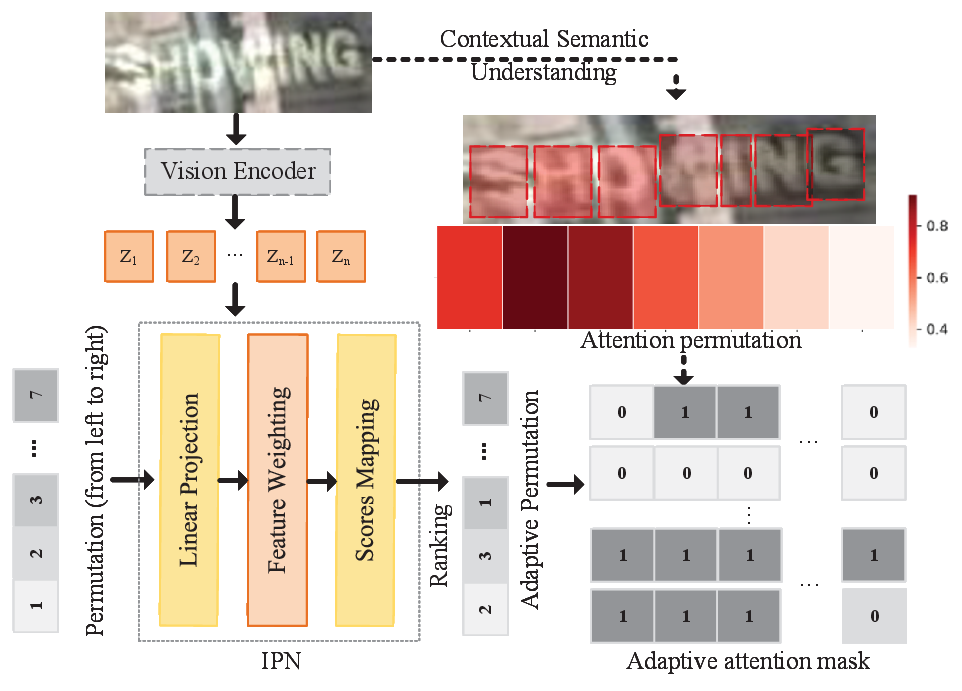}
\caption{Illustration of IPN. The solid line represents the process of mask generation i.e. non-linear weighted mapping of left-to-right permutations. The dashed line represents the interpretation of the mask generation: the visual information guides the adaptive mask to learn the inter-correlation of the contextual positions. \label{fig1}}
\end{figure}

In contrast to Optical Character Recognition (OCR) in documents with more homogeneous text attributes, STR poses significant challenges due to irregular text, including rotated, curved, blurred, and occluded characters.
STR is primarily a vision-based task that involves two different modalities: image and text. Early works use a model pre-trained on image data and directly classify characters according to visual information of the text region without incorporating context, i.e., the meaning of text \cite{shi2018aster, yu2020towards}. However, visual information alone is insufficient in cases where the text is unreadable, e.g., occluded, blurred, or with various fonts and spatial layouts \cite{bautista2022scene}. By language modeling (LM), text is introduced to aid recognition as text carries rich contextual information. Modeling the dependencies between the visual information and context and using autoregressive sequence modeling to predict character tokens \cite{wang2021two, fang2021read, na2022multi, wu2023str, bautista2022scene, zhao2023clip4str, xu2024ote}.

Current LM-based approaches can be divided into two categories: external LM-based \cite{wang2021two, fang2021read, na2022multi, wu2023str} and internal LM-based \cite{bautista2022scene, zhao2023clip4str, xu2024ote}. The external LM-based approach introduces an independent LM to correct the token prediction of the visual model by explicitly modeling the context. In contrast, the internal LM-based approach embeds LM in the visual model to exploit joint visual information and context.
External LM is used as a spell checker to rectify the visual prediction via unidirectional sequence modeling. The conditional independence of the LM from the input may cause it to correct the predictions erroneously \cite{fang2021read, wu2023str}. Therefore, the internal LM-based approach applies Permutation Language Modeling (PLM) to enhance the correlation between the input image and context, improving the generalization of sequence modeling \cite{bautista2022scene, zhao2023clip4str}. It can be seen as an ensemble of autoregressive models with shared architecture and weights \cite{bautista2022scene}. Using attention masks to dynamically specify token dependencies, PLM allows the decoder to perform sequence modeling of characters in arbitrary orders without relying on specific sequence order \cite{zhao2023clip4str}. Compared to unidirectional modeling, such as left-to-right (LTR) and right-to-left (RTL), the model can capture more information about token dependencies.



{However, there are still two problems: 1:) PLM produces a training fit oscillation problem due to the introduction of randomized permutation. Randomized permutation brings stochastic token dependencies that cause the model to have an unstable fitting process in decoding, which is defined as training fit oscillation in this paper. This makes decoding inefficient and requires multi-training to determine the best model. 2:) Iterative refinement (IR) was initially proposed in external LM-based approaches \cite{wang2021two, fang2021read, na2022multi, wu2023str} and then in internal LM-based ones \cite{bautista2022scene, zhao2023clip4str, xu2024ote} to aggregate global information from visual and language models to refine prediction. However, internal LM-based models have adopted an attention-based encoder-decoder architecture that integrates visual and language modeling. It means that contextual semantics can be refined from visual information in the decoder instead of using IR, which is inefficient and increases decoding time and complexity.
}


\begin{figure}[]
\centering
\includegraphics[width=8 cm]{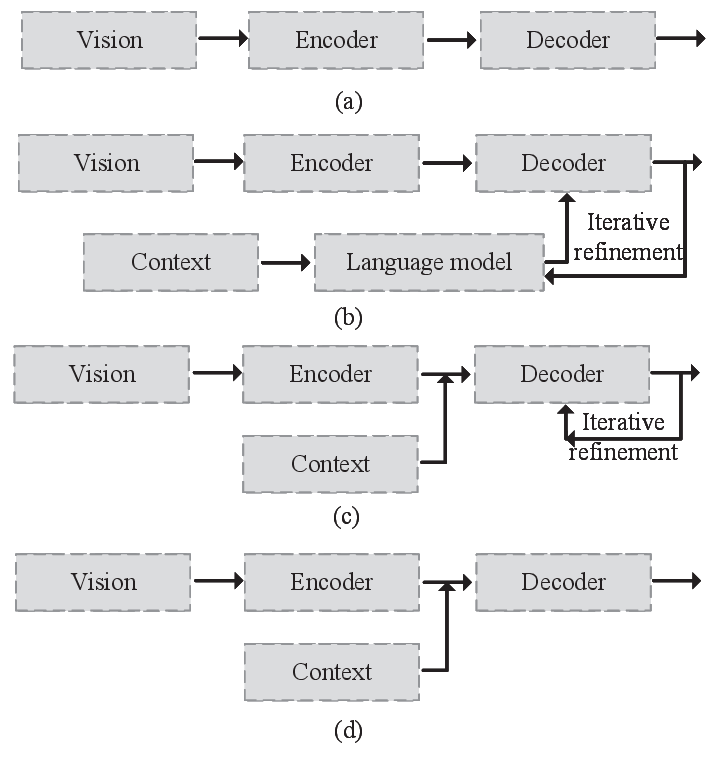}
\caption{The basic flow of STR. (a) Visual feature coding and decoding (b) Joint visual and context representation based on external LM. (c) Joint visual and context representation based on internal LM. (d) Internal LM-based visual-context representation without Iterative Refinement (IR) (Ours).\label{fig2}}
\end{figure}


To avoid the training fit oscillation and the use of IR, this paper proposes the Hierarchical Attention autoregressive model with Adaptive Permutation (HAAP) to improve the efficiency of cross-modal information decoding. In particular, first, we propose Implicit Permutation Neurons (IPN) to establish an adaptive attention mask in decoding by jointly considering the visual information and context (as illustrated in Fig. 1). IPN adaptively exploits token dependencies, enhancing the input-output correlations within the model and preventing the training fit oscillation.
Second, we develop the Cross-modal Hierarchical Attention mechanism (CHA) to integrate cross-modal information by establishing the dependencies among the position queries, context and the visual information using the adaptive attention mask provided by IPN. The contextual semantics are refined through a hidden layer, enabling each position token to be predicted based on global contextual semantics, rather than solely on preceding tokens. CHA significantly improves decoding efficiency and avoids the need for IR (as shown in Fig. 2).

The contributions of this paper can be summarized as follows:
\begin{itemize}
\item	This paper proposes an internal LM-based HAAP to end-to-end coupling of position, image, and context to improve the generalization ability of the model in various STR scenes.
\item	We propose a novel neuronal structure, i.e., IPN, to efficiently capture the token dependencies via an adaptive attention mask. The model automatically learns contextual semantic representations.
\item A hierarchical feature processing CHA is introduced to establish the position-context-image interaction. Enhancing the correlation between visual information and context and expanding the semantic receptive field of each position token.
\item	HAAP achieves state-of-the-art (SOTA) results on the STR benchmark for all character sets as well as on larger, more difficult real datasets that include occluded and arbitrarily oriented text {(as shown in Table. \ref{table-2})}. Besides, HAAP also shows the cost-quality trade-offs in terms of parameters, FLOPs, and runtime usage (as shown in Table. \ref{table-4}).
\end{itemize}

The rest of the paper is structured as follows: section II discusses related work. Section III describes the architecture of the proposed algorithm in detail. Section IV describes the experiments and analysis and the summary in Section V.

\section{Related work}
In this section, we review current deep learning-based STR methods. Recent studies are mainly categorized as context-independent and context-relevant. We summarize and discuss the research in these two directions separately.

\subsection{Context-independent STR}
Since STR is a visual task, the researchers intuitively represent the input using a visual model and perform the class prediction from visual features. The parallel prediction-based methods directly use Fully Convolutional Network (FCN) \cite{long2015fully} to segment characters pixel by pixel without considering the relationship between characters. Liao et al \cite{lyu2018mask} recognized characters by grouping segmented pixels into text regions. Wan et al \cite{wan2020textscanner} identified characters in the correct order using an additional segmentation map. Multi-class prediction cannot accurately construct a complete text instance because the output characters are conditionally uncorrelated.

Immediately, sequence modeling-based methods use Recurrent Convolutional Networks (RNN) to capture the correlation of characters \cite{liang2015recurrent}. The most typical is the Connectionist Temporal Classification (CTC)-based scheme \cite{graves2012connectionist}. The RNN models the feature sequence extracted by a convolutional neural network (CNN) and is end-to-end trained using CTC loss \cite{shi2016end}. The RNN receives the current input and the previous hidden state at each time step and outputs the current representation. This sequence modeling structure allows the RNN to naturally adapt to the characteristics of sequential data, to capture the linguistic rules and semantics in the text. RNN is also used in a series of Attention-based \cite{bahdanau2014neural} schemes. Several attempts have explored the enhancement of image encoding and decoding by employing different attention mechanisms \cite{cheng2018aon, lee2016recursive, li2019show}. Character classification by aligning features and character positions, or converting STR into a multi-instance learning problem \cite{yu2020towards, jaderberg2014synthetic}. Sequence modeling-based methods have made great progress with the introduction of the Transformer \cite{vaswani2017attention, dosovitskiy2020image}. Transformer was then proposed as an alternative to RNN for sequence modeling \cite{atienza2021vision, baek2021if, 9695247, 9765383, xue2023image, xia2023scene, diao2024hierarchical}. However, context-independent methods that rely only on image features for prediction cannot address recognition in low-quality images, especially being less robust to corruption such as occlusion, distortion, or incomplete characters. This limitation motivates using context to make recognition models more robust.

\subsection{Context-relevant STR}
Context-dependent STR is a typical cross-modal fusion process since the internal interactions between vision and context are constructed. Considering that contextual semantics can assist visual models in processing heterogeneous data, the community has recently appeared with a large number of research studies using LM to assist recognition \cite{qiao2020seed, wang2021two, fang2021read, bautista2022scene, na2022multi, zhang2022context, zhao2023clip4str, wu2023str}. VisionLAN \cite{wang2021two} introduced character masking to guide the visual model to refer to text information in the visual context. MATRN \cite{na2022multi} referenced context to enhance visual semantics based on spatial coding. STRT \cite{wu2023str} designed an iterative text Transformer to predict the probability distribution in character sequences. It is shown that robust recognition models can be constructed by introducing external LMs for explicit modeling.
Considering that character sequences are usually modeled in a left-to-right manner, the researcher develops the integrated model to capture twice the amount of information through the bi-directional LM \cite{yu2020towards, shi2018aster}. SRN \cite{yu2020towards} combined the features of two unidirectional Transformers for prediction, which resulted in twice as expensive both computationally and parametrically. Therefore, ABINet \cite{fang2021read} used a novel bidirectional completion network based on bidirectional feature representation. STRT \cite{wu2023str} used LM to capture the linguistic rule and refine the position token prediction. However, the conditional independence of external LM from the input image limits its performance and makes the model erroneously rectify the correct predictions. Subsequently, the internal LM was proposed for implicit language modeling to improve image-context interaction. PARSeq \cite{bautista2022scene} used PLM to enhance the generalization of internal LM. It supports flexible decoding by dynamically specifying the dependence of position tokens and then decoding the position tokens from the input. CLIP4STR \cite{zhao2023clip4str} introduced CLIP \cite{radford2021learning} and used PLM to construct an encoder-decoder framework to integrate image and text information.

PLM-based methods mitigate the problem of noisy inputs and increase the diversity of the data by dynamically specifying the sequence order. However, the use of randomized permutations introduces the problem of training fit oscillation.
Furthermore, most internal LM-based models employ an attention-based encoder-decoder architecture to integrate visual information with contextual semantics. Nevertheless, due to inefficient decoding, these models typically adopt IR to refine prediction, at the cost of increased computational complexity. Randomized token dependencies make an insufficient contextual semantic representation. Since each position token prediction only relies on the semantics preceding the tokens, the model needs to use IR further to aggregate global contextual semantics. This paper proposes to adopt an adaptive sequence modeling strategy that promotes training data diversity and effectively captures the interdependence between visual and contextual information. To further enhance the efficiency of cross-modal information decoding, the Multi-head attention mechanism (MHA) is used to hierarchically model relationships among position queries, contextual semantics, and visual information. The correlation between position and context is enhanced, enabling each token to access global contextual semantics and avoiding the need for IR.

\begin{figure*}[]
\centering
\includegraphics[width=18 cm]{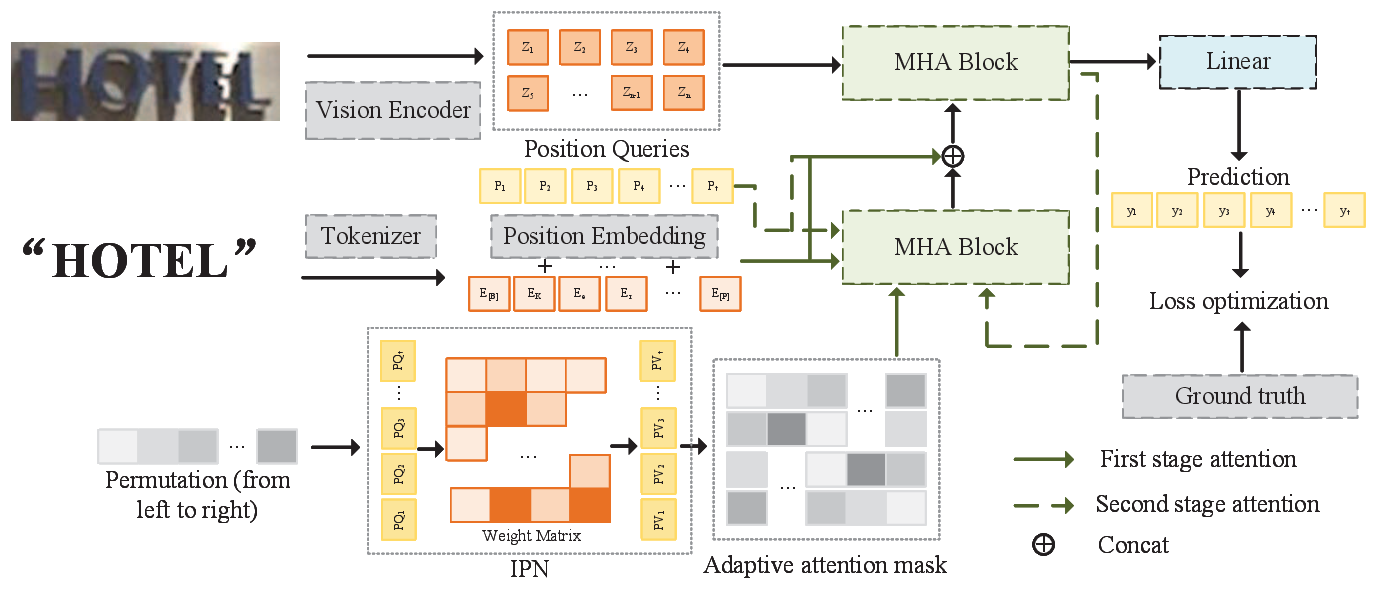}
\caption{The pipeline of HAAP. (a) Image and text inputs are represented as a series of patches and semantic tokens. (b) Visual information guides the IPN to assign bidirectional and adaptive masks to the context. (c) MHA is used to perform hierarchical image-context interaction and decoding.\label{fig3}}
\end{figure*}

\section{Methodology}
In this section, we present in detail the principles and framework of our proposed Hierarchical Attention autoregressive Model with Adaptive Permutation (HAAP), including Implicit Permutation Neurons (IPN) and Cross-modal Hierarchical Attention mechanism (CHA).

\subsection{Overview}
HAAP follows an encoder-decoder architecture as shown in Fig. \ref{fig3}. In the encoding phase, the image and text inputs are represented as a series of patches and context tokens. Subsequently, MHA is used for hierarchical image-context interaction and decoding. Specifically, first, the Transformer encoder \cite{dosovitskiy2020image} is used to establish the internal feature representation and correlation of the image patches. Second, the IPN assigns bi-directional and adaptive attention masks to the context for adjusting the order of autoregressive sequence modeling. The first attention stage establishes the dependencies between the visual information and context with attention masks, producing the contextual semantics. The second attention stage encodes the contextual semantics and position through MHA into position queries. Finally, guide with visual information, position queries autoregressive decode the position token per iteration.

\begin{figure}[]
\centering
\includegraphics[width=4.5 cm]{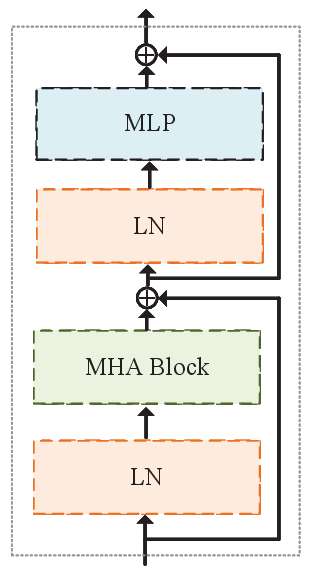}
\caption{Illustration of a ViT layer from Dosovitskiy et al. \cite{dosovitskiy2020image}. $LN$ pertains to layer normalization. $MLP$ represents Multilayer Perceptron.\label{fig4}}
\end{figure}

\subsection{Encoder}

\textbf{Vision encoder}
As shown in Fig. \ref{fig4}, HAAP uses $L = 12$ layers of the Transformer encoder \cite{dosovitskiy2020image} to build the visual encoder. The output of the last encoder is processed by layer normalization $LN$. First, the image of size $H \times W \times C$ (height $H$, width $W$, channels $C$) is reshaped into a series of flat 2D patches $x_p \in  \mathbb{R} ^{N \times (P^2 - C)}$, where $(P, P)$ is the resolution of each image block and $N = HW/P^2$ is the number of blocks generated. The patch $x_p$ is flattened and mapped to $D$ dimensions using a trainable linear projection $E \in \mathbb{R} ^{({P^2}C)\times D}$. The position embedding $E_{pos} \in \mathbb{R} ^{(N+1) \times D}$ is added to the projected output $z$, as:
\begin{align}
z = [x^0_pE;x^1_pE;x^2_pE;\cdots;x^N_pE;] + E_{pos},
\label{eq1}
\end{align}
where $x^i_p$ represents the $i$-th $x_p$. $z$ performs feature extraction in alternating MHA and Mult-Layer Perceptron (MLP). $LN$ is applied initially in each block to mitigate internal covariate shift during training. Additionally, residual concatenation is applied twice in the block to aid gradient propagation in the depth model, as:
\begin{align}
z^{\prime}_l = MHA(LN(z_{l-1})) + z_{l-1}, l\in [1,L]
\label{eq2}
\end{align}
\begin{align}
z_l = MLP(LN(z^{\prime}_l)) + z^{\prime}_l, l\in [1, L]
\label{eq3}
\end{align}
where $MLP$ contains two nonlinear layers with GELU activation functions. The $z^{\prime}_l$ and $z_l$ are encoded visual information. MHA is the extension of scaled dot-product attention to multiple representation subspaces or heads \cite{bautista2022scene}.

\textbf{Context encoder} The context uses a Tokenizer \cite{bautista2022scene} for lowercase byte pair encoding BPE. The start, padding, and end of the text sequence are padded with [SOS], [PAD] and [EOS] tokens, respectively.

\subsection{Implicit Permutation Neurons (IPN)}
Considering that STR traditionally relies on left-to-right $P_{left2right}$ or right-to-left $P_{right2left}$ unidirectional sequence modeling, we propose IPN to enhance the correlation between inputs and outputs through multidirectional sequence modeling. Essentially, multidirectional sequence modeling considers internal correlations of context while keeping attention to the bi-directional information flow to capture the word language rules and avoid the effort of visual interference (i.g, occlusion and distortion). For example, to predict the first ``e" in the word ``clearance," we consider ``cl\_," ``ecnara\_," and the context of ``cl\_ar". IPN provides an attention mask during the attention operation to generate dependencies between the input context and the output without actually replacing the text labels. {Table. \ref{table-1} illustrates three examples of the mask.}

\begin{table}[]
\centering
\caption{Illustration of Masks from IPN. $P_{left2right}$ and $P_{right2left}$ represent the left-to-right and right-to-left permutation.[B] And [E] represent the beginning and end of the sequence. $y1,y2,y3$ represent the three positions of this sequence. 1 indicates that the output has a conditional dependency on the input. 0 indicates that there is no information flowing from input to output.\label{table-1}}
\begin{tabular}{ccccc}
\hline
\multicolumn{5}{c}{$P_{left2right}$ Mask}        \\ \hline
         & {[}B{]} & y1 & y2 & y3 \\
y1       & 1       & 0  & 0  & 0  \\
y2       & 1       & 1  & 0  & 0  \\
y3       & 1       & 1  & 1  & 0  \\
{[}E{]}  & 1       & 1  & 1  & 1  \\ \hline
\multicolumn{5}{c}{$P_{right2left}$ Mask}        \\ \hline
         & {[}B{]} & y1 & y2 & y3 \\
y1       & 1       & 0  & 1  & 1  \\
y2       & 1       & 0  & 0  & 1  \\
y3       & 1       & 0  & 0  & 0  \\
{[}E{]}  & 1       & 1  & 1  & 1  \\ \hline
\multicolumn{5}{c}{Attention mask of IPN} \\ \hline
         & {[}B{]} & y1 & y2 & y3 \\
y1       & 1       & 0  & 1  & 1  \\
y2       & 1       & 1  & 0  & 0  \\
y3       & 1       & 1  & 1  & 0  \\
{[}E{]}  & 1       & 1  & 1  & 1  \\ \hline
\end{tabular}
\end{table}

Formally, first, given the $T$-length text labels $y = [y^1, y^2, \cdots, y^T]$, decompose the likelihood $[1, 2, \cdots, T]$ according to the canonical order using the chain rule to build a left-to-right permutation of $P_{left2right}$ to obtain the model inference as:

\begin{equation}
\mathrm{log} p(y|x) = \sum_{t=1}^{T}{\mathrm{log} p_\theta(y_t|y_{<t}, x)},
\label{eq7}
\end{equation}
where  $\theta$ represents the model parameter and $x$ is the input image. Second, to learn and extract complex structural properties from the data, $P_{query}$ is assigned to $T$ positions in the $P_{left2right}$ sequence and projected to a high-dimensional space, as:
\begin{equation}
M_t = P_{left2right}^\intercal \cdot P_{query},
\label{eq8}
\end{equation}
where $M_t \in \mathbb{R}_T^T$ represents the intermediate matrix.

Third, the model learns the optimal representation without manually designing features via a learnable weight matrix $P_{weight}$, as:

\begin{equation}
M_w = M_t \cdot P_{weight},
\label{eq9}
\end{equation}
where $M_w$ is the weighted transformation matrix. This process helps the model capture the input representation. Even though there is no explicit nonlinear activation function throughout the process, the model can still capture the nonlinear relationships in the data because of the learning of the weights. The transformation matrix is then mapped to the sequence of scores $P_{score}$ via $P_{value}$, as:

\begin{equation}
P_{score} = M_w \cdot P_{value}^\intercal.
\label{eq10}
\end{equation}

Next, $P_{score}$ is normalized from largest to smallest to generate the actual ranking order. Immediately, we invert the left-to-right $[1, 2, . , T]$ and adaptive permutation to obtain twice the amount of information.

\subsection{Cross-modal Hierarchical Attention mechanism (CHA)}
In multi-directional sequence modeling, the standard Transformer-based autoregressive model is inefficient since the feature distribution predicted by the hidden state is independent of the target position \cite{yang2019xlnet}. Let us assume that we parameterize the next token distribution as:

\begin{equation}
p_\theta (x_{rt} = x | x_{r<t}) = \frac{exp(e(x)^\intercal h_\theta (x_{r<t}))}{\sum{x^\prime}exp(e(x \prime)^\intercal h_\theta (x_{r<t}))},
\label{eq15}
\end{equation}
where $h_\theta (x_{r<t})$ denotes the hidden representation of $x_{r<t}$ produced by the shared transformer network after proper masking. We use $x_{rt}$ and $x_{r<t}$ to denote the $t$-th and the first $t-1$ elements of the permutation set $r$. Since the same model parameter $\theta$ is shared between all decomposition sequences during training, $x_{rt}$ is expected to see every possible element in the sequence. Note that the position token $x_{rt}$ to be predicted only depends on the representation $h_\theta (x_{r<t} )$. If only using the initial contextual semantics learned from the multi-directional sequence modeling without considering the current input visual information, it is not possible to update the representation $h_\theta (x_{r<t} )$ with global information, increasing the probability of error accumulation. For this, researchers employ IR to improve the representation of $h_\theta (x_{r<t})$ \cite{fang2021read, bautista2022scene}. The IR uses the cloze mask to enhance the understanding of $h_\theta (x_{r<t})$ for the context because the cloze mask is able to provide the rest of the information other than the current position.

To avoid this problem, we introduce CHA to hierarchically encode context and position by allowing context, position, and image to interact with each other through a two-stream attention mechanism. We use two sets of hidden representations, i.e., from contextual semantic and query representations. In the first stage, context-image attention is established, producing the context semantic representation, which encodes both the context and $x_{rt}$ itself tokens guided by visual information, as:

\begin{equation}
Attn_{cv} = Attn_c + dropout[MHA(Attn_c,z_l)],
\label{eq11}
\end{equation}
where
\begin{equation}
Attn_c = c + dropout[MHA(c,c,M)],
\label{eq12}
\end{equation}
where $c \in \mathbb{R}^{(T +1) \times d_{model}}$ is the context embeddings with position information and $M \in \mathbb{R}^{(T +1)\times (T +1)}$ is the attention mask. The total sequence length is increased to $T + 1$ because of the use of special delimiter tokens ([B] or [E]).
Next, MHA is used to establish position-context-image attention. as:

\begin{equation}
Attn_f = Attn_p +dropout[MHA(Attn_p,z_l)],
\label{eq13}
\end{equation}
where
\begin{equation}
Attn_p = p + dropout[MHA(p, Attn_{cv},M)],
\label{eq13}
\end{equation}
where $p$ is the position queries. By updating the contextual semantic representation, each position token can gather the global information guided by the current input.
CHA first helps the model capture dependencies between context and the visual information by allowing sequence interactions with the adaptive attention mask ($M$) provided by IPN. Then, the correlation between position and context is enhanced through context-image implicit guidance, thus getting rid of additional IR. The robustness of the model is improved because the uniform gradient flow allows IPN and CHA to interact with each other during training.
Finally, the output logits $y^{\prime}$ are obtained as:

\begin{equation}
y^{\prime} = Linear(Attn_f) \in R^{(T + 1)×(S+1)},
\label{eq16}
\end{equation}
where $S$ is the size of the character set used for training. Additional characters are associated with the [E] token (the end of the token sequence). In the optimization phase, the training loss is the average of the  $K$ cross-entropy loss $\mathcal{L}_{ce}$ with attention mask, as:

\begin{equation}
\mathcal{L} = \frac{1}{k} \sum_{k=1}^{K} {\mathcal{L}_{ce}(y^{\prime}_k, y)},
\label{eq14}
\end{equation}
where $y^{\prime}_k$ represents the $k$-th output logit and $y$ is the real label. $K=4$ is the number of permutations.

\section{Experiments}

\subsection{Dataset}
This experiment uses both synthetic training datasets MJSynth (MJ) \cite{jaderberg2014synthetic} and SynthText (ST) \cite{gupta2016synthetic} as well as real data for training including nine real-world datasets (COCO-Text (COCO) \cite{veit2016coco}, RCTW17 \cite{shi2017icdar2017}, Uber-Text (Uber) \cite{zhang2017uber}, ArT \cite{chng2019icdar2019}, LSVT \cite{sun2019icdar}, MLT19 \cite{nayef2019icdar2019}, ReCTS \cite{zhang2019icdar}, TextOCR \cite{singh2021textocr} and OpenVINO \cite{krylov2021open}). We use IIIT 5k-word (IIIT5k) \cite{mishra2012scene}, CUTE80 (CUTE) \cite{risnumawan2014robust}, Street View Text (SVT) \cite{wang2011end}, SVT-Perspective (SVTP) \cite{phan2013recognizing}, ICDAR 2013 (IC13) \cite{karatzas2013icdar} and ICDAR 2015 (IC15) \cite{karatzas2015icdar}, COCO, Uber, ArT for testing. The dataset contains challenging texts such as multi-orientation, curved, blurry, low resolution, distorted, and occluded text.

\textbf{Synthetic datasets}
MJ consists of 8.9 million photorealistic text images composed of synthetic datasets through over 90k English dictionaries. The composition of MJ consists of three parts background, foreground, and optional shadow/border respectively. It uses 1400 different fonts. Additionally, the font word spacing, thickness, underlining, and other attributes of MJ are different. MJ also utilizes different background effects, border/shadow rendering, base shading, projection distortion, natural image blending, and noise. ST consists of 8 million word text images synthesized by blending text over natural images. It uses scene geometry, textures, and surface normals to naturally blend and distort text renderings on the surfaces of objects in the image. Similar to MJ, ST uses randomized fonts for its text. The text image is cropped from the natural image in which the synthesized text is embedded.

\textbf{Real-world datasets}
The COCO dataset has a total of 73,127 training and 9.8k test images containing non-text, legible, illegible, and occluded text images. Each image contains at least one instance of readable text. RCTW contains 12,263 annotated images of the large-scale Chinese field dataset. The Uber dataset has 127,920 training 60,000 test images collected from Bing Maps Streetside. It contains house numbers and vertical and rotated text on signboards. The ArT dataset was created to recognize arbitrarily shaped text containing 32,028 training images and 26,000 test images of arbitrarily shaped, perspective, rotated, or curved text. SVT is a large-scale street scene text dataset collected from streets in China, including 41,439 text images of arbitrarily shaped natural scenes. MLT19 was created to recognize multilingual text. It has a total of 56,727 training images consisting of seven languages: Arabic, Latin, Chinese, Japanese, Korean, Bengali and Hindi. The ReCTS dataset contains 26,432 irregular texts arranged in various layouts or written in unique fonts. TextOCR and OpenVINO are large datasets with diverse images containing 818,087 and 2,071,541 images, respectively. They have complex scenes with multiple objects and text of different resolutions, orientations, and qualities.

The IIIT5k dataset is a collection of 5000 natural scenes and digitally born text images crawled from Google image search, containing 2,000 images for evaluation and 3,000 images for testing. It has low resolution, multiple font styles, light and dark transformations, and projection distortion. CUTE contains 288 cropped images for the curved text collection. The images were captured by digital cameras or collected from the Internet. SVT is a collection of street view text from Google Street View. It contains 257 images for evaluation and 647 images for testing. Further, the SVTP dataset is a more complex collection of images containing numerous perspective texts, totaling 645 images for testing. IC13 and IC15 were created for the ICDAR Robust Reading Contest. IC13 contains 848 images for training and 1,015 images for evaluation. IC15 contains 4,468 images for training and 2,077 images for evaluation. Many of the samples contain perspective and blurred text.

\begin{table*}[]
\centering
\caption{The comparison of word accuracy on six common benchmarks in terms of both context relevant and independent. ``S" and ``R" denote Synthetic and real data, respectively.\label{table-2}}
\begin{tabular}{cccccccccc}
\hline
Method                                                 & Train data & IIIT5K(\%)    & SVT(\%)       & IC13(\%)      & IC13(\%)      & IC15(\%)      & IC15(\%)      & SVTP(\%)      & CUTE(\%)      \\ \hline
\multicolumn{10}{l}{Context-independent}                                                                                                                                                            \\ \hline
CRNN(2016)\cite{shi2016end}           & S          & 84.3          & 78.9          & –             & 88.8          & –             & 61.5          & 64.8          & 61.3          \\
SRN(2020)\cite{yu2020towards}         & S          & 94.8          & 91.5          & 95.5          & –             & 82.7          & –             & 85.1          & 87.8          \\
ViTSTR(2021)\cite{atienza2021vision}  & S          & 88.4          & 87.7          & 93.2          & 92.4          & 78.5          & 72.6          & 81.8          & 81.3          \\
TRBA(2021)\cite{baek2021if}           & S          & 92.1          & 88.9          & –             & 93.1          & –             & 74.7          & 79.5          & 78.2          \\
TRAN(2023)\cite{9695247}            & S          & 94.3          & 90.7          &               & 93.6          &               & 77.7          & 85.0          & 81.6          \\
DRNet(2023)\cite{9765383}          & S          & 93.4          & 90.6          &               & 95.8          &               & 81.6          & 83.6          & 83.0          \\
CASR-DRNet(2023)\cite{9623473}          & S          & 95.1          & 92.3          &               & 95.3          &               & 83.0          & 85.0          & 89.2          \\
I2C2W(2023)\cite{xue2023image}        & S          & 94.3          & 91.7          &               & 95.0          &               & 82.8          & 83.7          & 93.1          \\
Xia et. al.(2023)\cite{xia2023scene}  & S          & 91.9          & 91.3          & 96.1          & 94.2          & 80.6          & 78.0          & 83.4          & 88.8          \\
HVSI(2024)\cite{diao2024hierarchical} & S          & 96.7          & 94.9          &               & 97.4          &               & 86.2          & 89.2          & 95.5          \\
CAM(2024)\cite{yang2024class} & S          & 97.4          & 96.1          &               & 97.2          &               & 87.8          & 90.6          & 92.4          \\
\multicolumn{10}{l}{Context-relevant}                                                                                                                                                               \\ \hline
SEED(2020)\cite{qiao2020seed}         & S          & 93.8          & 89.6          &               & 92.8          &               & 80.0          & 81.4          & 83.6          \\
VisionLAN(2021)\cite{wang2021two}     & S          & 95.8          & 91.7          &               & 95.7          &               & 83.7          & 86.0          & 88.5          \\
ABINet(2021)\cite{fang2021read}       & S          & 96.2          & 93.5          & 97.4          & –             & 86.0          & –             & 89.3          & 89.2          \\
ABINet(2021)\cite{fang2021read}       & R          & 98.6          & \textbf{98.2} & 98.0          & 97.8          & 90.5          & 88.7          & 94.1          & 97.2          \\
ConCLR(2022)\cite{zhang2022context}   & S          & 96.5          & 94.3          &               & 97.7          &               & 85.4          & 89.3          & 91.3          \\
MATRN(2022)\cite{na2022multi}         & S          & 96.6          & 95.0          & 97.9          & 95.8          & 86.6          & 82.8          & 90.6          & 93.5          \\
PARSeq(2022)\cite{bautista2022scene}  & S          & 97.0          & 93.6          & 97.0          & 96.2          & 86.5          & 82.9          & 88.9          & 92.2          \\
PARSeq(2022)\cite{bautista2022scene}  & R          & 99.1          & 97.9          & 98.3          & 98.4          & 90.7          & 89.6          & 95.7          & 98.3          \\
STRT(2023)\cite{wu2023str}            & S          & 97.6          & 95.7          &               & 97.6          &               & 86.7          & 90.1          & 94.9          \\
OTE(2024)\cite{xu2024ote}            & S          & 96.2          & 93.5          & 97.6              &           & 85.9              &           & 89.6          & 91.7          \\
HAAP                                                   & S          & 97.7          & 95.1          & 97.2          & 96.4          & 87.8          & 85.6          & 90.7          & 95.1          \\
HAAP                                                   & R          & \textbf{99.3} & \textbf{98.2} & \textbf{98.7} & \textbf{98.8} & \textbf{91.3} & \textbf{90.7} & \textbf{97.4} & \textbf{99.3} \\ \hline
\end{tabular}
\end{table*}

\subsection{Implementation Details}
The entire training and testing is implemented in two RTX 3090 GPUs with 24GB of RAM. The training is performed with a batch size of 1024 for 31,815 iterations i.e. 10 epochs on a real dataset of 3,257,585 samples and 4 epochs on a synthetic dataset of 16.89M samples. Adam \cite{lan2012optimal} optimizer is used along with the 1cycle \cite{smith2019super} learning rate scheduler with an initial learning rate of $7e-5$. We set a maximum label length of $T = 25$ and used a character set of size $S = 94$ containing mixed-case alphanumeric characters and punctuation. The image is augmented using a three-layer RandAugment \cite{cubuk2020randaugment} operation where sharpening is tuned to Gaussian Blur and Poisson Noise and resized to $128 \times 32$ pixels. Each patch is set a size of $8 \times 4$. For the model testing phase, we use a 36-character set i.e. containing numbers and 26 letters. We use word accuracy as the main evaluation metric i.e., the prediction is considered correct if and only if the characters match at all positions.

\textbf{Remark} The highlighted red and red lines in the visualization results indicate error prediction and lack of word, respectively. The best performance result is shown in bold font.

\subsection{Comparison with State-of-the-Arts}
We compare HAAP with recent methods on 6 public benchmarks, {and the quantitative results are listed in Table. \ref{table-2}}.  We can see that HAAP outperforms the benchmark Parseq \cite{bautista2022scene}, realizing SOTA performance.  The recognition performances in IC15 (incident scene text), SVTP (perspective scene text), and CUTE80 (curved text) are improved by $1.1\%$, $1.7\%$, and $1\%$ respectively. Our model achieves $99.3\%$ recognition accuracy on the IIIT5K (distorted, low-resolution scene text) dataset. This is an improvement of $2.6\%$ compared to the latest method, HVSI \cite{diao2024hierarchical}. This verifies the excellent performance of HAAP on irregular text datasets, which contain a large number of low-quality images such as distorted and blurred images.

\begin{figure*}[]
\centering
\includegraphics[width=14 cm]{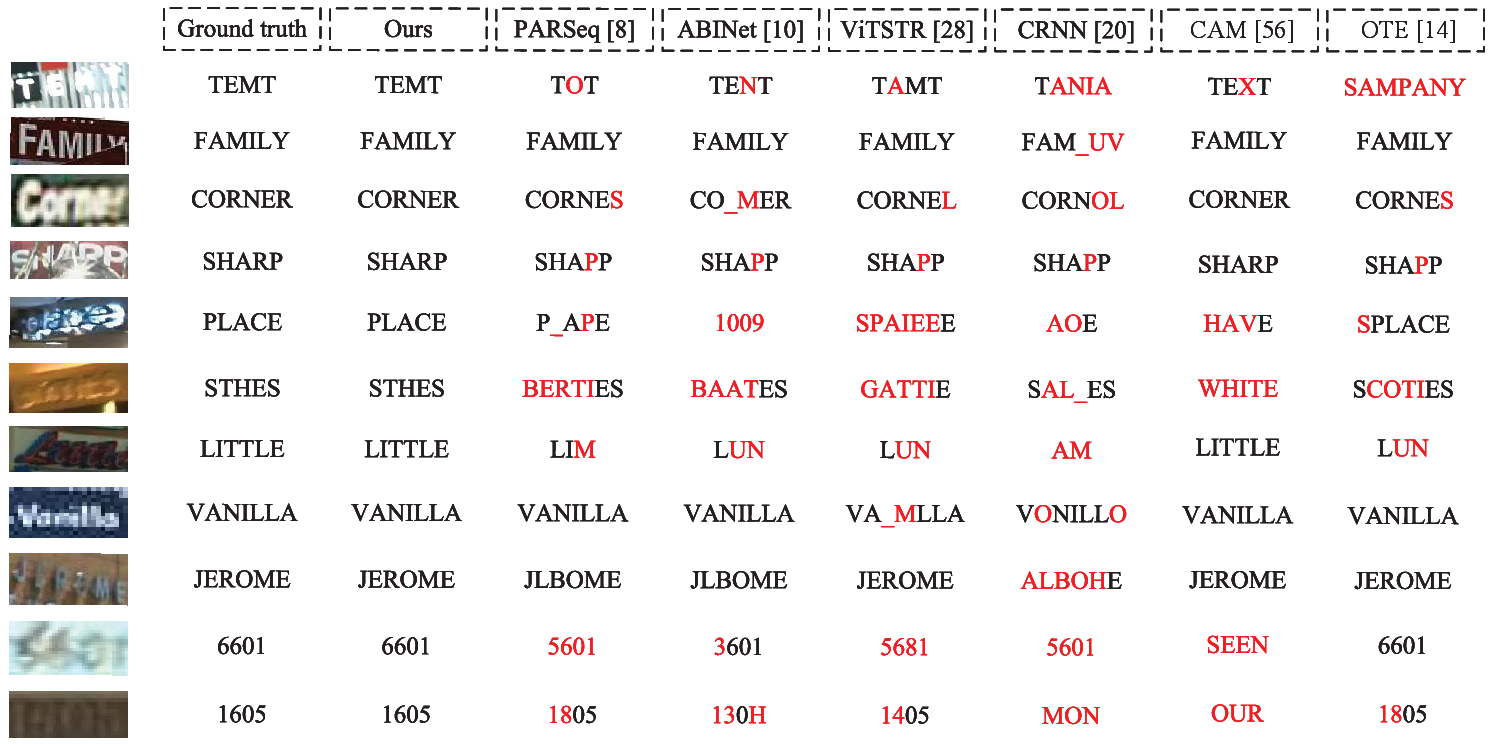}
\caption{Comparison of qualitative results on challenging textual data including blurring, distortion, occlusion, low resolution, perspective-shifting, and multi-directionality. \label{fig6}}
\end{figure*}

Besides validating the model on small-scale public benchmarks, we evaluate HAAP on three larger and more challenging recent benchmarks consisting of irregular text with different shapes, low-resolution, rotated, and occluded text. {The results are shown in Table. \ref{table-3}.} It achieves $80.6\%$, $85.5\%$, and $85.8\%$ outperforms the previous methods in terms of accuracy, respectively. {Representative visible results are shown in Fig. \ref{fig6}}, which demonstrates that our method is sufficiently robust to occlusion and text direction variability. In detail, we find that HAAP and recent methods are effective in solving complex background problems caused by perspective shifts or uneven lighting. However, artistic lettering and occlusion can lead to errors in previous work. The independence of heterogeneous data leads to ambiguous judgments by the recognizer, which can be effectively avoided by HAAP. Furthermore, HAAP can protect against additional distortions and ambiguities. {For example, although the second half of "CORNER" in Fig. \ref{fig6} is ambiguous}, our model can still reason effectively. {HAAP also recognizes that the "R" in "SHARP" in Fig. \ref{fig6}} relies on AR reasoning with adaptive permutations.

\begin{figure*}[]
\centering
\includegraphics[width=15 cm]{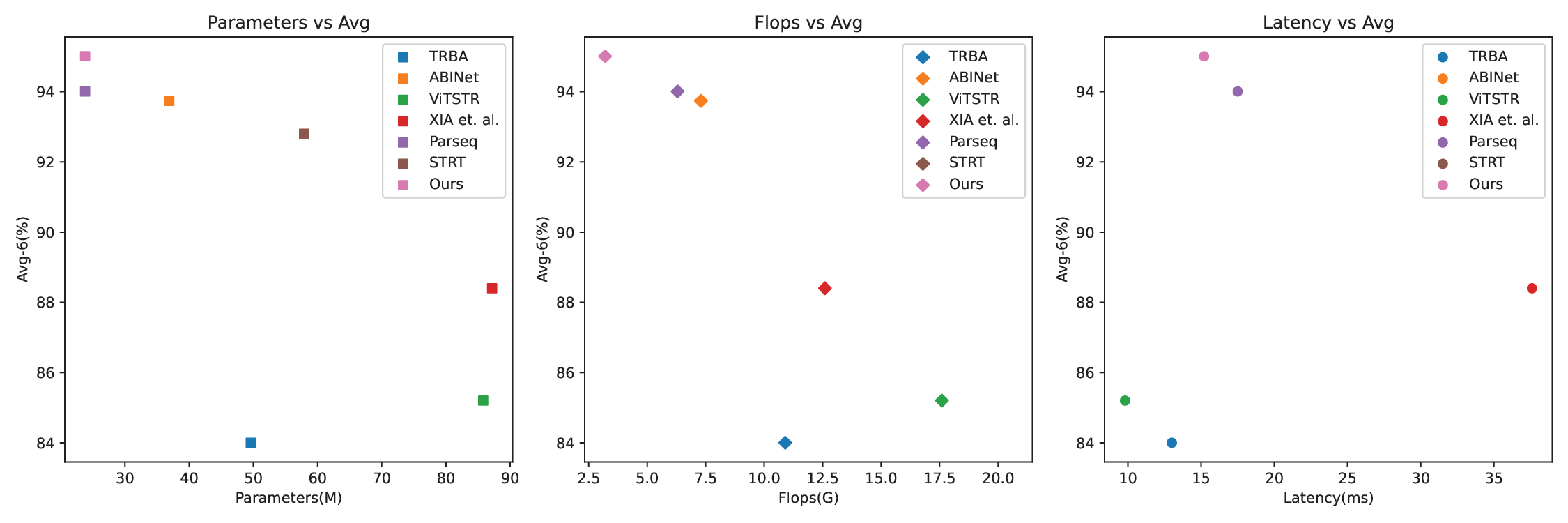}
\caption{Visual comparison with other methods in terms of complexity and efficiency. \label{fig5}}
\end{figure*}

\begin{table}[H]
\centering
\caption{The comparison of word accuracy on three challenging datasets. ``S" and ``R" denote Synthetic and real data, respectively.\label{table-3}}
\resizebox{0.4\textwidth}{!}{
\begin{tabular}{ccccc}
\hline
Method      & Train data & COCO          & ArT           & Uber          \\ \hline
CRNN(2016)\cite{shi2016end}    & S          & 49.3          & 57.3          & 33.1          \\
TRBA(2021)\cite{baek2021if}   & S          & 61.4          & 68.2          & 38.0          \\
ViTSTR(2021)\cite{atienza2021vision} & S          & 56.4          & 66.1          & 37.6          \\
ABINet(2021)\cite{fang2021read} & S          & 57.1          & 65.4          & 34.9          \\
PARSeq(2022)\cite{bautista2022scene} & S          & 64.0          & 70.7          & 42.0          \\
HAAP        & S          & \textbf{65.0}          & \textbf{71.7}          & \textbf{43.2}          \\ \hline
CRNN(2016)\cite{shi2016end}   & R          & 62.2          & 66.8          & 51.0          \\
TRBA(2021)\cite{baek2021if}   & R          & 77.5          & 82.5          & 81.2          \\
ViTSTR(2021)\cite{atienza2021vision} & R          & 73.6          & 81.0          & 78.2          \\
ABINet(2021)\cite{fang2021read} & R          & 76.5          & 81.2          & 71.2          \\
PARSeq(2022)\cite{bautista2022scene} & R          & 79.8          & 84.5          & 84.1          \\
HAAP        & R          & \textbf{80.6} & \textbf{85.5} & \textbf{85.8} \\ \hline
\end{tabular}}
\end{table}

\begin{table}[H]
\centering
\caption{Comparison with other methods in terms of complexity and efficiency. $AVG-6$ represents the average accuracy in the six benchmarks.\label{table-4}}
\begin{tabular}{lllll}
\hline
Method      & Parameters(M) & Flops(G) & Latency(ms) & Avg-6(\%) \\ \hline
TRBA\cite{baek2021if}        & 49.6          & 10.9     & 13.0        & 84.0    \\
ABINet\cite{fang2021read}      & 36.9          & 7.3      & 33.9        & 93.7    \\
ViTSTR\cite{atienza2021vision}      & 85.8          & 17.6     & 9.8         & 85.2    \\
XIA et. al.\cite{xia2023scene} & 87.2          & 12.6     & 37.6        & 88.4    \\
PARSeq\cite{bautista2022scene}      & 23.8          & 4.3      & 17.5        & 94.0    \\
STRT\cite{wu2023str}        & 57.9          & 20.6     & 31.8        & 92.8    \\
Ours        & \textbf{20.9}          & \textbf{3.6}      & \textbf{15.2}        & \textbf{95.0}    \\ \hline
\end{tabular}
\end{table}

In addition, {Fig. \ref{fig5} and Table. \ref{table-4} shows the trade-off between accuracy and cost (parameters, FLOPS, and latency).} For concise representation, we use $Avg$ to denote the average accuracy, which is computed as a weighted average based on the number of datasets. HAAP achieves the highest average word accuracy and exhibits competitive cost-quality in all three metrics. Compared to PARSeq, HAAP uses far fewer parameters and FLOPs. In terms of latency, {HAAP achieves $Avg-6$ of 95.0\%  with 15.2 ms/image}, which outperforms previous methods.

\begin{table}[H]
\centering
\caption{Ablation study of the IPN.$Avg-9$ represents the average accuracy in nine datasets. $P_{left2right}$ and $P_{right2left}$ represent the left-to-right and right-to-left permutation. \label{table-5}}
\begin{tabular}{lllll}
\hline
$P_{left2right}$ & $P_{right2left}$ & PLM & IPN & Avg-9(\%)        \\ \hline
\checkmark            & -            &  -   &  -  & 86.63          \\
\checkmark            & \checkmark            &-     &  -   & 89.99          \\
       -      &   -           & \checkmark   &  -  & 90.12          \\
        -     &   -           &  -   & \checkmark   & \textbf{90.40} \\ \hline
\end{tabular}
\end{table}

\subsection{Ablation study}
To evaluate the effectiveness of the proposed HAAP with IPN and CHA, we conduct ablation studies on nine benchmark datasets. In these experiments, all models are trained using real datasets with consistent settings of their model hyperparameters.

\begin{figure}[H]
	\centering
	\subfigure[the model with PLM] {\includegraphics[width=.45\textwidth]{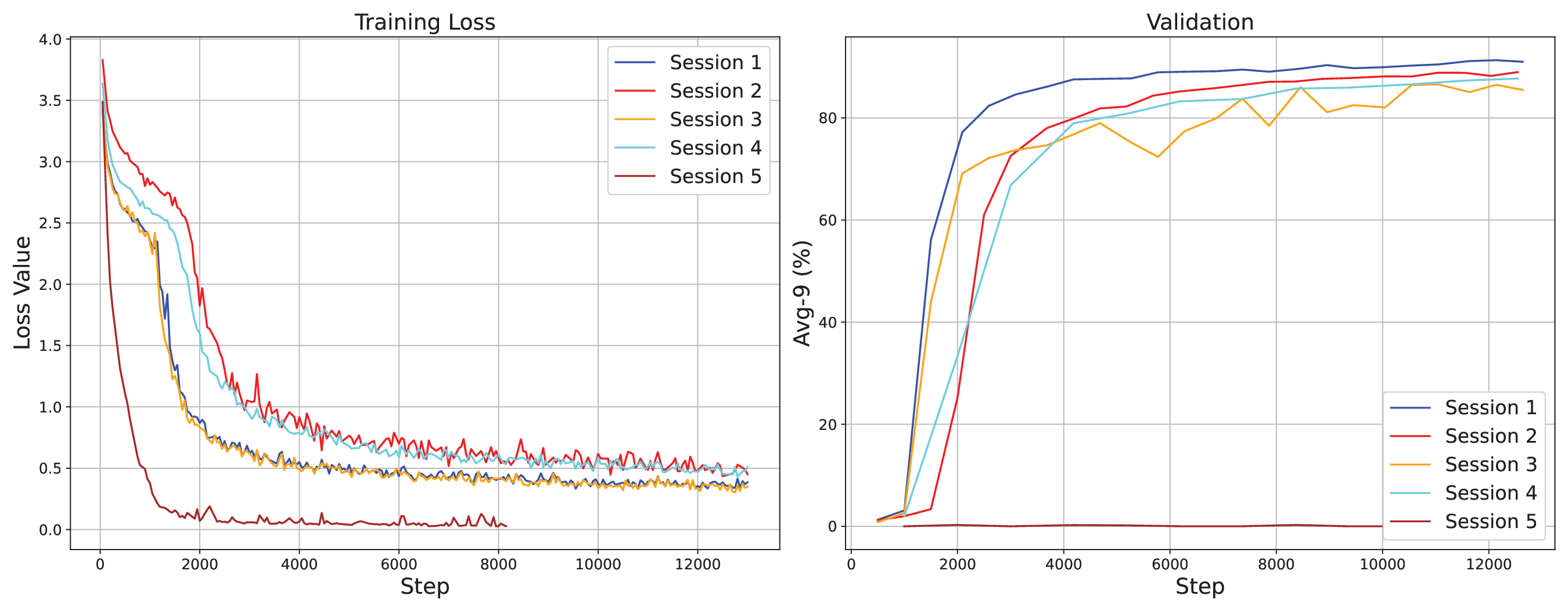}}
	\subfigure[the model with IPN] {\includegraphics[width=.45\textwidth]{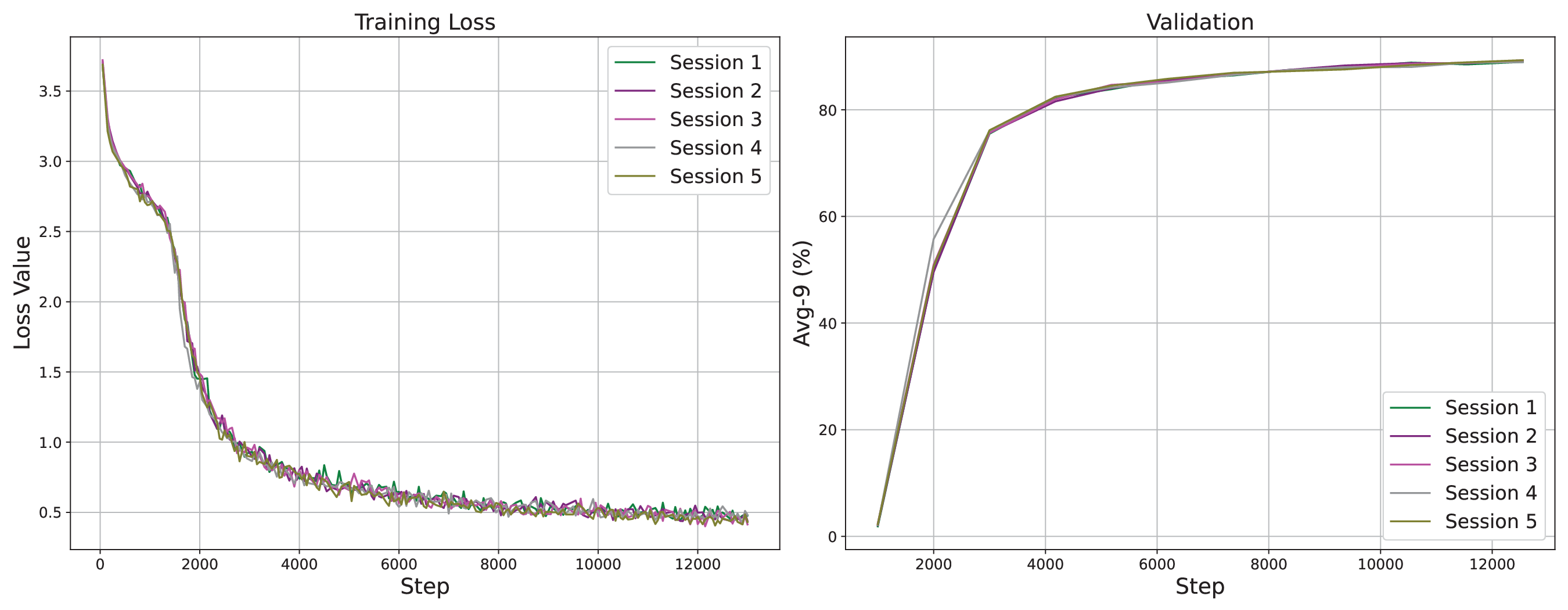}}
	\caption{Comparison of Loss (Left) and Valuation (Right) curves in five continuous training sessions.}
	\label{fig7}
\end{figure}

\begin{figure}[H]
\centering
\includegraphics[width=8 cm]{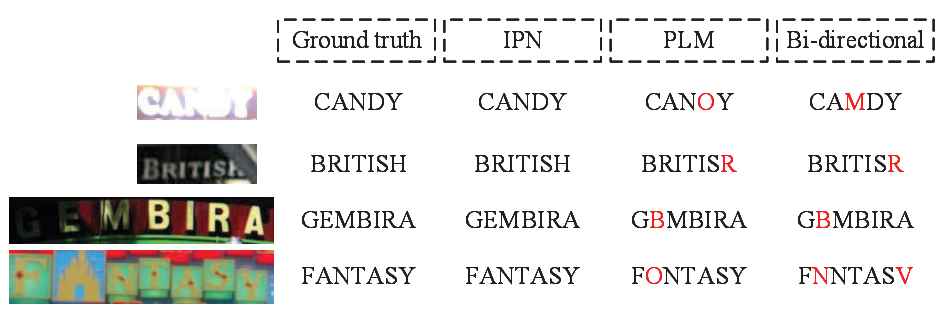}
\caption{The qualitative results of the IPN. \label{fig8}}
\end{figure}

\textbf{The effectiveness of IPN}
{We compare IPN with different sequence modeling principles (i.g, left-to-right, bi-directional, and PLM) to evaluate the effectiveness of adaptive sequence modeling. Table. \ref{table-5} shows that the bidirectional model can achieve a performance gain of about $3.4\%$ on $Avg-9$ compared to the left-to-right model.  Compared to the PLM model ($90.12\%$), the IPN model achieves a performance gain of $0.28\%$ in terms of $Avg-9$. To demonstrate the effectiveness of IPN in model training, we show the training and validation curves in five continuous training sessions in Fig. \ref{fig7}. Meanwhile, we use early stopping to determine the iteration step of model convergence by validating that the increase of $Avg-9$ is less than $0.3\%$ in 2000 steps. The result is depicted in Tab. \ref{table-IPN_convergence}. Specifically, as shown in Fig. \ref{fig7}, the IPN model can achieve more stable training and convergence than the PLM model in multi-training.  As depicted in Tab. \ref{table-IPN_convergence}, the $Avg-9$ standard deviation (Std Dev) of the IPN model is $0.33$, which is much less than the PLM model ($1.45$). With IPN, adaptive sequence modeling can help the model avoid train fit oscillation and achieve convergence advantage with less training data and time costs than PLM. The qualitative results in Fig. \ref{fig8} show that the IPN model can effectively predict some complex cases influenced by background.}


\begin{table}[H]
\centering
\caption{Training schedule comparison in five continuous training sessions. Std Dev represents standard deviation. \label{table-IPN_convergence}}
\resizebox{0.45\textwidth}{!}{
\begin{tabular}{ccccc}
\hline
Training session & Step of convergence    & Avg-9 (\%) & Sample count (M) & Training hours \\ \hline
PLM                &         &            &                  &                \\
Session 1          & 15318   & 90.12      & 15.69            & 5.23           \\
Session 2          & 19591   & 88.89      & 20.06            & 5.79           \\
Session 3          & 18500   & 86.83      & 18.94            & 5.64           \\
Session 4          & 20091   & 87.62      & 20.57            & 5.86           \\
Session 5          & 999     & 0          & 10.23            & 2.6            \\ \hline
Mean               & 18375   & 88.37      & 18.82            & 5.63           \\
Std Dev            & 2143.53 & 1.45       & 2.19             & 0.28           \\ \hline
IPN                &         &            &                  &                \\
Session 1          & 11636   & 90.37      & 11.92            & 4.67           \\
Session 2          & 12545   & 89.67      & 12.85            & 4.79           \\
Session 3          & 12545   & 89.76      & 12.85            & 4.79           \\
Session 4          & 13727   & 89.78      & 14.06            & 4.91           \\
Session 5          & 12545   & 90.31      & 12.85            & 4.79           \\ \hline
Mean               & \textbf{12599}   & \textbf{89.98}      & \textbf{12.91}            & \textbf{4.79}           \\
Std Dev            & \textbf{743.05}  & \textbf{0.33}       & \textbf{0.76}             & \textbf{0.08}           \\ \hline
\end{tabular}}
\end{table}

\begin{figure}[H]
\centering
\includegraphics[width=8 cm]{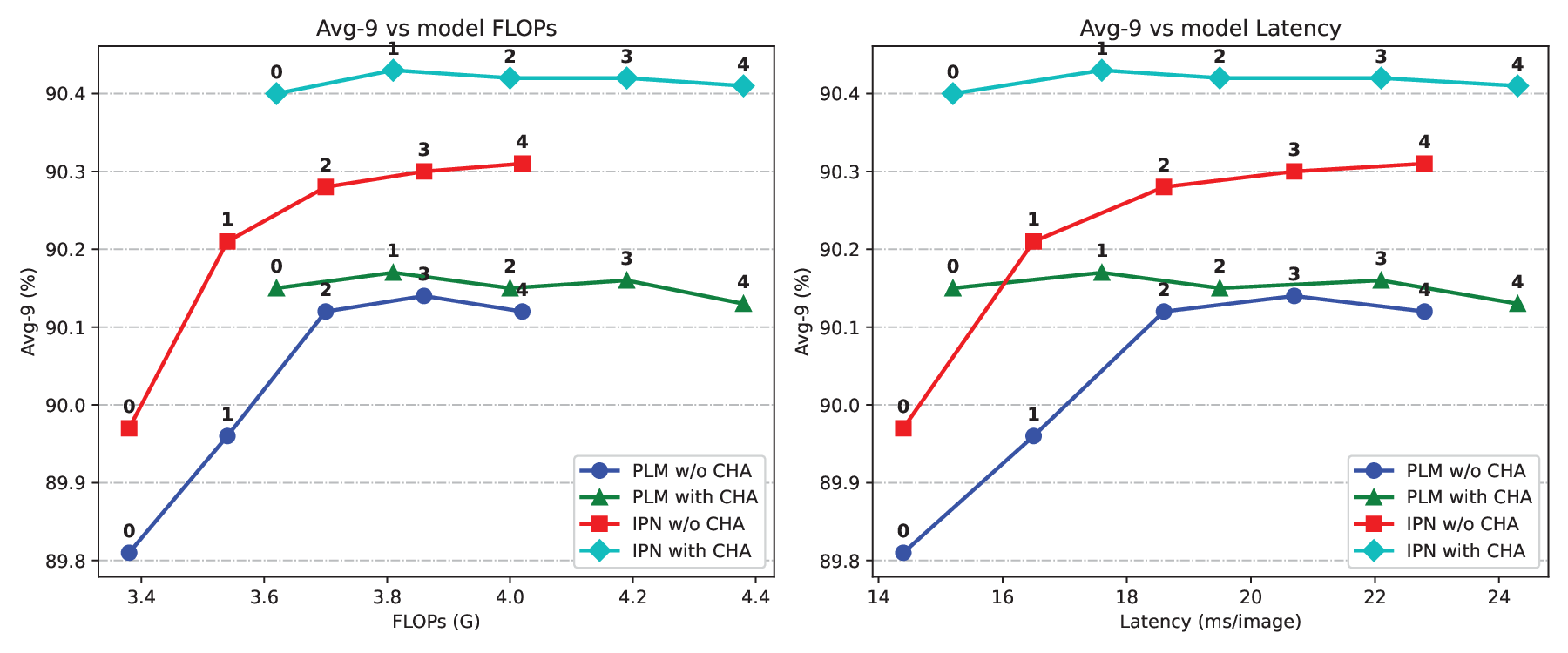}
\caption{Average accuracy (Avg-9) in nine datasets and FLOPs of the model (Left) and single-image latency (Right) for each decoding scheme. The number of IR used is indicated for each point. \label{fig_total}}
\end{figure}

\textbf{The effectiveness of CHA}
{To evaluate the effectiveness of CHA, we compare PLM-based and IPN-based models with CHA in different numbers of IRs. As shown in Table. \ref{table-6}, we find that both the PLM model and IPN model without CHA require additional IR to improve the autoregressive generalization. By using CHA, the PLM model achieves the $Avg-9$ of $90.15\%$ without IR, outperforming the one after iterating IR three times without  CHA ($90.14\%$). Based on CHA, the IPN model achieves a performance gain of $0.43\%$ without IR, indicating the effectiveness of CHA in decoding position tokens.
In addition, Fig. \ref{fig_total} shows the trade-off between the $Avg-9$, the model computational complexity (FLOPs), and single-image latency (Latency) with the number of IRs. The   model without CHA requires two times of IR to achieve the stable performance, while the one with CHA maintains the performance advantage regardless of IR. Thus, the use of   CHA avoids the additional computational complexity and latency induced by IR.}

\begin{figure*}[]
\centering
\includegraphics[width=15 cm]{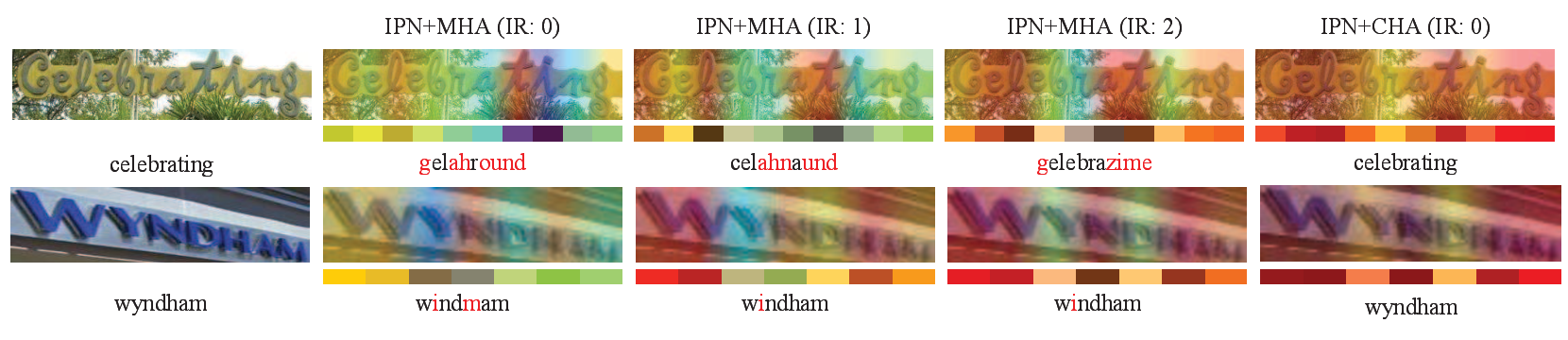}
\caption{Comparison of confidence heat maps. Red color indicates the highest score, and blue color indicates the lowest score. For each approach from top to bottom are the score region visualization, confidence level, and prediction results. Ground truths are shown below the original images. \label{fig-ir}}
\end{figure*}

\begin{figure}[H]
\centering
\includegraphics[width=7 cm]{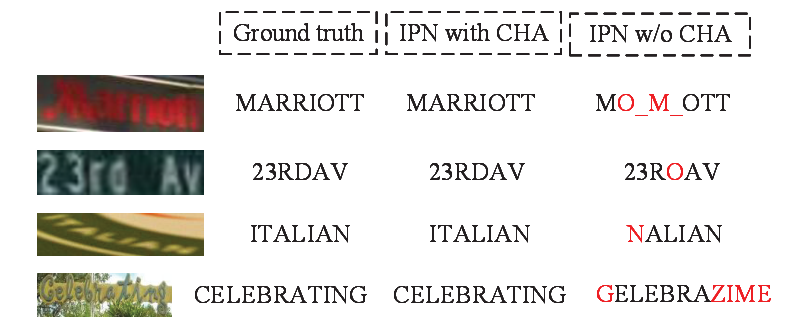}
\caption{Comparison of qualitative results.\label{fig9}}
\end{figure}

As shown in Fig. \ref{fig-ir}, although IR enhances the confidence of the IPN model, it results in wrong predictions. In contrast, the IPN model with CHA can generate high-confidence and accurate predictions without IR. The results clearly show that CHA can improve the performance of the autoregressive model through position-context-image interactions without the need for additional contextual semantic aggregation through IR. The qualitative results in Fig. \ref{fig9} show that the CHA-based model effectively recognizes low-resolution and blurred texts.


\begin{table}[H]
\centering
\caption{The ablation study of CHA. \label{table-6}}
\begin{tabular}{cccccc}
\hline
\multirow{2}{*}{\begin{tabular}[c]{@{}c@{}}Algorithm \\ and complexity\end{tabular}} & \multicolumn{5}{c}{Number of IR (Avg-9.(\%))}                                      \\ \cline{2-6}
                                                                                     & 0              & 1              & 2              & 3              & 4              \\ \hline
PLM w/o CHA                                                                              & 89.81          & 89.96          & 90.12          & 90.14          & 90.12          \\
IPN w/o CHA                                                                              & 89.97          & 90.21          & 90.28          & 90.30          & 90.31          \\
FLOPs(G)                                                                             & 3.38           & 3.54           & 3.70           & 3.86           & 4.02           \\ \hline
PLM with CHA                                                                              & 90.15          & 90.17          & 90.15          & 90.16          & 90.13          \\
IPN with CHA                                                                              & \textbf{90.40} & \textbf{90.43} & \textbf{90.42} & \textbf{90.42} & \textbf{90.41} \\
FLOPs(G)                                                                             & 3.62           & 3.81           & 4.00           & 4.19           & 4.38           \\ \hline
\end{tabular}
\end{table}




\subsection{Limitation and discussion}
\begin{figure}[H]
\centering
\includegraphics[width=8 cm]{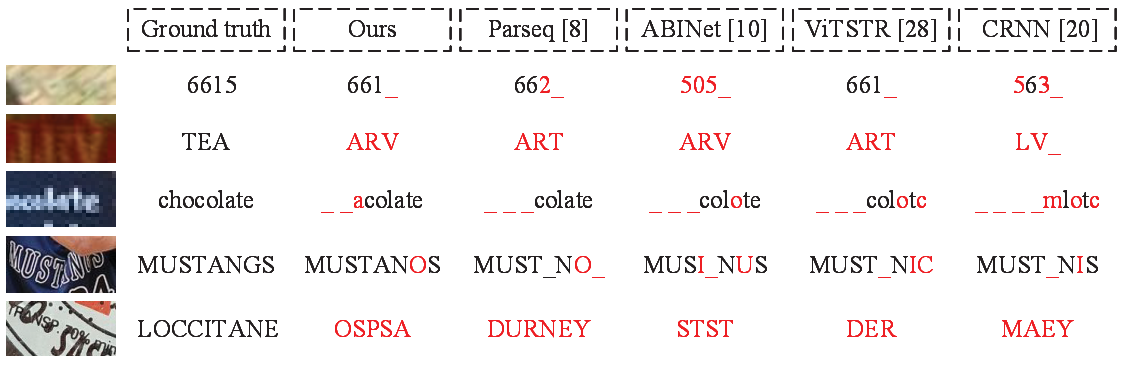}
\caption{Visualization results of some failure cases.\label{fig10}}
\end{figure}

Although the proposed method can accurately recognize arbitrary-shaped texts in most cases, it still struggles with four kinds of tiny texts (the total number of pixels does not exceed 2000) due to the lack of visual representation of tiny text.: i:) target confused with the background; ii:) blurred long-text; iii:) deformed text under multi-occlusion; and iv:) overlapping text. Fig. \ref{fig10} shows examples of such failure cases.

\section{Conclusion}
This paper proposes HAAP to enhance the position-context-image interaction capability using IPN and CHA to improve autoregressive generalization. First, IPN captures the correlation between context and visual information by dynamically specifying token dependencies with an adaptive attention mask. Adaptive sequence modeling increases the diversity of the training data and prevents the model from training fit oscillation. Second, CHA module for hierarchical feature processing is proposed to exploit position-context-image semantic dependencies in the sequence without IR. The experimental results validate the effectiveness and SOTA performance of HAAP in terms of accuracy, complexity, and latency. In the future, we expect to extend HAAP to end-to-end detection and recognition, where the detector can be embedded in the visual decoder since position awareness is exhibited using the attention mechanism.


%



\section*{Acknowledgment}
This work was supported by the NSFC (No. 62171135), Fujian Key Project (No.2023XQ004), Fujian Eagle-Scholar and Distinguished Project (No. 2022J06010), Fujian Natural Science Fund (No. 2024H6013).

\ifCLASSOPTIONcaptionsoff
  \newpage
\fi



\small
\bibliographystyle{IEEEtran}
%
\bibliography{refer}

@article{song2022clip,
  title={Clip models are few-shot learners: Empirical studies on vqa and visual entailment},
  author={Song, Haoyu and Dong, Li and Zhang, Wei-Nan and Liu, Ting and Wei, Furu},
  journal={arXiv preprint arXiv:2203.07190},
  year={2022}
}

@incollection{taki2023scene,
  title={Scene Text Recognition for Text-Based Traffic Signs},
  author={Taki, Youssef and Zemmouri, Elmoukhtar},
  booktitle={Advances in Intelligent Traffic and Transportation Systems},
  pages={67--77},
  year={2023},
  publisher={IOS Press}
}

@article{luo2022clip4clip,
  title={Clip4clip: An empirical study of clip for end to end video clip retrieval and captioning},
  author={Luo, Huaishao and Ji, Lei and Zhong, Ming and Chen, Yang and Lei, Wen and Duan, Nan and Li, Tianrui},
  journal={Neurocomputing},
  volume={508},
  pages={293--304},
  year={2022},
  publisher={Elsevier}
}

@inproceedings{qiao2020seed,
  title={Seed: Semantics enhanced encoder-decoder framework for scene text recognition},
  author={Qiao, Zhi and Zhou, Yu and Yang, Dongbao and Zhou, Yucan and Wang, Weiping},
  booktitle={Proceedings of the IEEE/CVF conference on computer vision and pattern recognition},
  pages={13528--13537},
  year={2020}
}

@article{vaswani2017attention,
  title={Attention is all you need},
  author={Vaswani, Ashish and Shazeer, Noam and Parmar, Niki and Uszkoreit, Jakob and Jones, Llion and Gomez, Aidan N and Kaiser, {\L}ukasz and Polosukhin, Illia},
  journal={Advances in neural information processing systems},
  volume={30},
  year={2017}
}

@article{shi2018aster,
  title={Aster: An attentional scene text recognizer with flexible rectification},
  author={Shi, Baoguang and Yang, Mingkun and Wang, Xinggang and Lyu, Pengyuan and Yao, Cong and Bai, Xiang},
  journal={IEEE transactions on pattern analysis and machine intelligence},
  volume={41},
  number={9},
  pages={2035--2048},
  year={2018},
  publisher={IEEE}
}

@inproceedings{yu2020towards,
  title={Towards accurate scene text recognition with semantic reasoning networks},
  author={Yu, Deli and Li, Xuan and Zhang, Chengquan and Liu, Tao and Han, Junyu and Liu, Jingtuo and Ding, Errui},
  booktitle={Proceedings of the IEEE/CVF conference on computer vision and pattern recognition},
  pages={12113--12122},
  year={2020}
}

@inproceedings{wang2021two,
  title={From two to one: A new scene text recognizer with visual language modeling network},
  author={Wang, Yuxin and Xie, Hongtao and Fang, Shancheng and Wang, Jing and Zhu, Shenggao and Zhang, Yongdong},
  booktitle={Proceedings of the IEEE/CVF International Conference on Computer Vision},
  pages={14194--14203},
  year={2021}
}

@inproceedings{fang2021read,
  title={Read like humans: Autonomous, bidirectional and iterative language modeling for scene text recognition},
  author={Fang, Shancheng and Xie, Hongtao and Wang, Yuxin and Mao, Zhendong and Zhang, Yongdong},
  booktitle={Proceedings of the IEEE/CVF Conference on Computer Vision and Pattern Recognition},
  pages={7098--7107},
  year={2021}
}

@inproceedings{na2022multi,
  title={Multi-modal text recognition networks: Interactive enhancements between visual and semantic features},
  author={Na, Byeonghu and Kim, Yoonsik and Park, Sungrae},
  booktitle={European Conference on Computer Vision},
  pages={446--463},
  year={2022},
  organization={Springer}
}

@article{wu2023str,
  title={STR transformer: a cross-domain transformer for scene text recognition},
  author={Wu, Xing and Tang, Bin and Zhao, Ming and Wang, Jianjia and Guo, Yike},
  journal={Applied Intelligence},
  volume={53},
  number={3},
  pages={3444--3458},
  year={2023},
  publisher={Springer}
}

@inproceedings{bautista2022scene,
  title={Scene text recognition with permuted autoregressive sequence models},
  author={Bautista, Darwin and Atienza, Rowel},
  booktitle={European Conference on Computer Vision},
  pages={178--196},
  year={2022},
  organization={Springer}
}

@article{zhao2023clip4str,
  title={CLIP4STR: A Simple Baseline for Scene Text Recognition with Pre-trained Vision-Language Model},
  author={Zhao, Shuai and Wang, Xiaohan and Zhu, Linchao and Yang, Yi},
  journal={arXiv preprint arXiv:2305.14014},
  year={2023}
}

@inproceedings{long2015fully,
  title={Fully convolutional networks for semantic segmentation},
  author={Long, Jonathan and Shelhamer, Evan and Darrell, Trevor},
  booktitle={Proceedings of the IEEE conference on computer vision and pattern recognition},
  pages={3431--3440},
  year={2015}
}

@inproceedings{lyu2018mask,
  title={Mask textspotter: An end-to-end trainable neural network for spotting text with arbitrary shapes},
  author={Lyu, Pengyuan and Liao, Minghui and Yao, Cong and Wu, Wenhao and Bai, Xiang},
  booktitle={Proceedings of the European conference on computer vision (ECCV)},
  pages={67--83},
  year={2018}
}

@inproceedings{wan2020textscanner,
  title={Textscanner: Reading characters in order for robust scene text recognition},
  author={Wan, Zhaoyi and He, Minghang and Chen, Haoran and Bai, Xiang and Yao, Cong},
  booktitle={Proceedings of the AAAI conference on artificial intelligence},
  volume={34},
  number={07},
  pages={12120--12127},
  year={2020}
}

@inproceedings{liang2015recurrent,
  title={Recurrent convolutional neural network for object recognition},
  author={Liang, Ming and Hu, Xiaolin},
  booktitle={Proceedings of the IEEE conference on computer vision and pattern recognition},
  pages={3367--3375},
  year={2015}
}

@article{graves2012connectionist,
  title={Connectionist temporal classification},
  author={Graves, Alex and Graves, Alex},
  journal={Supervised sequence labelling with recurrent neural networks},
  pages={61--93},
  year={2012},
  publisher={Springer}
}

@article{shi2016end,
  title={An end-to-end trainable neural network for image-based sequence recognition and its application to scene text recognition},
  author={Shi, Baoguang and Bai, Xiang and Yao, Cong},
  journal={IEEE transactions on pattern analysis and machine intelligence},
  volume={39},
  number={11},
  pages={2298--2304},
  year={2016},
  publisher={IEEE}
}

@article{bahdanau2014neural,
  title={Neural machine translation by jointly learning to align and translate},
  author={Bahdanau, Dzmitry and Cho, Kyunghyun and Bengio, Yoshua},
  journal={arXiv preprint arXiv:1409.0473},
  year={2014}
}

@inproceedings{cheng2018aon,
  title={Aon: Towards arbitrarily-oriented text recognition},
  author={Cheng, Zhanzhan and Xu, Yangliu and Bai, Fan and Niu, Yi and Pu, Shiliang and Zhou, Shuigeng},
  booktitle={Proceedings of the IEEE conference on computer vision and pattern recognition},
  pages={5571--5579},
  year={2018}
}

@inproceedings{lee2016recursive,
  title={Recursive recurrent nets with attention modeling for ocr in the wild},
  author={Lee, Chen-Yu and Osindero, Simon},
  booktitle={Proceedings of the IEEE conference on computer vision and pattern recognition},
  pages={2231--2239},
  year={2016}
}

@inproceedings{li2019show,
  title={Show, attend and read: A simple and strong baseline for irregular text recognition},
  author={Li, Hui and Wang, Peng and Shen, Chunhua and Zhang, Guyu},
  booktitle={Proceedings of the AAAI conference on artificial intelligence},
  volume={33},
  number={01},
  pages={8610--8617},
  year={2019}
}

@article{jaderberg2014synthetic,
  title={Synthetic data and artificial neural networks for natural scene text recognition},
  author={Jaderberg, Max and Simonyan, Karen and Vedaldi, Andrea and Zisserman, Andrew},
  journal={arXiv preprint arXiv:1406.2227},
  year={2014}
}

@inproceedings{atienza2021vision,
  title={Vision transformer for fast and efficient scene text recognition},
  author={Atienza, Rowel},
  booktitle={International Conference on Document Analysis and Recognition},
  pages={319--334},
  year={2021},
  organization={Springer}
}

@inproceedings{baek2021if,
  title={What if we only use real datasets for scene text recognition? toward scene text recognition with fewer labels},
  author={Baek, Jeonghun and Matsui, Yusuke and Aizawa, Kiyoharu},
  booktitle={Proceedings of the IEEE/CVF Conference on Computer Vision and Pattern Recognition},
  pages={3113--3122},
  year={2021}
}

@ARTICLE{9695247,
  author={Wu, Lintai and Xu, Yong and Hou, Junhui and Chen, C. L. Philip and Liu, Cheng-Lin},
  journal={IEEE Transactions on Multimedia}, 
  title={A Two-Level Rectification Attention Network for Scene Text Recognition}, 
  year={2023},
  volume={25},
  number={},
  pages={2404-2414},
  doi={10.1109/TMM.2022.3146779}}

@ARTICLE{9765383,
  author={Li, Ming and Fu, Bin and Chen, Han and He, Junjun and Qiao, Yu},
  journal={IEEE Transactions on Multimedia}, 
  title={Dual Relation Network for Scene Text Recognition}, 
  year={2023},
  volume={25},
  number={},
  pages={4094-4107},
  doi={10.1109/TMM.2022.3171108}}

@article{xue2023image,
  title={Image-to-Character-to-Word Transformers for Accurate Scene Text Recognition},
  author={Xue, Chuhui and Huang, Jiaxing and Zhang, Wenqing and Lu, Shijian and Wang, Changhu and Bai, Song},
  journal={IEEE Transactions on Pattern Analysis and Machine Intelligence},
  year={2023},
  publisher={IEEE}
}

@article{xia2023scene,
  title={Scene text recognition based on two-stage attention and multi-branch feature fusion module},
  author={Xia, Shifeng and Kou, Jinqiao and Liu, Ningzhong and Yin, Tianxiang},
  journal={Applied Intelligence},
  volume={53},
  number={11},
  pages={14219--14232},
  year={2023},
  publisher={Springer}
}

@article{diao2024hierarchical,
  title={Hierarchical visual-semantic interaction for scene text recognition},
  author={Diao, Liang and Tang, Xin and Wang, Jun and Xie, Guotong and Hu, Junlin},
  journal={Information Fusion},
  volume={102},
  pages={102080},
  year={2024},
  publisher={Elsevier}
}

@article{dosovitskiy2020image,
  title={An image is worth 16x16 words: Transformers for image recognition at scale},
  author={Dosovitskiy, Alexey and Beyer, Lucas and Kolesnikov, Alexander and Weissenborn, Dirk and Zhai, Xiaohua and Unterthiner, Thomas and Dehghani, Mostafa and Minderer, Matthias and Heigold, Georg and Gelly, Sylvain and others},
  journal={arXiv preprint arXiv:2010.11929},
  year={2020}
}

@inproceedings{zhang2022context,
  title={Context-based contrastive learning for scene text recognition},
  author={Zhang, Xinyun and Zhu, Binwu and Yao, Xufeng and Sun, Qi and Li, Ruiyu and Yu, Bei},
  booktitle={Proceedings of the AAAI Conference on Artificial Intelligence},
  volume={36},
  number={3},
  pages={3353--3361},
  year={2022}
}

@inproceedings{radford2021learning,
  title={Learning transferable visual models from natural language supervision},
  author={Radford, Alec and Kim, Jong Wook and Hallacy, Chris and Ramesh, Aditya and Goh, Gabriel and Agarwal, Sandhini and Sastry, Girish and Askell, Amanda and Mishkin, Pamela and Clark, Jack and others},
  booktitle={International conference on machine learning},
  pages={8748--8763},
  year={2021},
  organization={PMLR}
}

@inproceedings{gupta2016synthetic,
  title={Synthetic data for text localisation in natural images},
  author={Gupta, Ankush and Vedaldi, Andrea and Zisserman, Andrew},
  booktitle={Proceedings of the IEEE conference on computer vision and pattern recognition},
  pages={2315--2324},
  year={2016}
}

@article{veit2016coco,
  title={Coco-text: Dataset and benchmark for text detection and recognition in natural images},
  author={Veit, Andreas and Matera, Tomas and Neumann, Lukas and Matas, Jiri and Belongie, Serge},
  journal={arXiv preprint arXiv:1601.07140},
  year={2016}
}

@inproceedings{shi2017icdar2017,
  title={Icdar2017 competition on reading chinese text in the wild (rctw-17)},
  author={Shi, Baoguang and Yao, Cong and Liao, Minghui and Yang, Mingkun and Xu, Pei and Cui, Linyan and Belongie, Serge and Lu, Shijian and Bai, Xiang},
  booktitle={2017 14th iapr international conference on document analysis and recognition (ICDAR)},
  volume={1},
  pages={1429--1434},
  year={2017},
  organization={IEEE}
}

@inproceedings{zhang2017uber,
  title={Uber-text: A large-scale dataset for optical character recognition from street-level imagery},
  author={Zhang, Ying and Gueguen, Lionel and Zharkov, Ilya and Zhang, Peter and Seifert, Keith and Kadlec, Ben},
  booktitle={SUNw: Scene Understanding Workshop-CVPR},
  volume={2017},
  pages={5},
  year={2017}
}

@inproceedings{chng2019icdar2019,
  title={Icdar2019 robust reading challenge on arbitrary-shaped text-rrc-art},
  author={Chng, Chee Kheng and Liu, Yuliang and Sun, Yipeng and Ng, Chun Chet and Luo, Canjie and Ni, Zihan and Fang, ChuanMing and Zhang, Shuaitao and Han, Junyu and Ding, Errui and others},
  booktitle={2019 International Conference on Document Analysis and Recognition (ICDAR)},
  pages={1571--1576},
  year={2019},
  organization={IEEE}
}

@inproceedings{sun2019icdar,
  title={ICDAR 2019 competition on large-scale street view text with partial labeling-RRC-LSVT},
  author={Sun, Yipeng and Ni, Zihan and Chng, Chee-Kheng and Liu, Yuliang and Luo, Canjie and Ng, Chun Chet and Han, Junyu and Ding, Errui and Liu, Jingtuo and Karatzas, Dimosthenis and others},
  booktitle={2019 International Conference on Document Analysis and Recognition (ICDAR)},
  pages={1557--1562},
  year={2019},
  organization={IEEE}
}

@inproceedings{nayef2019icdar2019,
  title={ICDAR2019 robust reading challenge on multi-lingual scene text detection and recognition—RRC-MLT-2019},
  author={Nayef, Nibal and Patel, Yash and Busta, Michal and Chowdhury, Pinaki Nath and Karatzas, Dimosthenis and Khlif, Wafa and Matas, Jiri and Pal, Umapada and Burie, Jean-Christophe and Liu, Cheng-lin and others},
  booktitle={2019 International conference on document analysis and recognition (ICDAR)},
  pages={1582--1587},
  year={2019},
  organization={IEEE}
}

@inproceedings{zhang2019icdar,
  title={Icdar 2019 robust reading challenge on reading chinese text on signboard},
  author={Zhang, Rui and Zhou, Yongsheng and Jiang, Qianyi and Song, Qi and Li, Nan and Zhou, Kai and Wang, Lei and Wang, Dong and Liao, Minghui and Yang, Mingkun and others},
  booktitle={2019 international conference on document analysis and recognition (ICDAR)},
  pages={1577--1581},
  year={2019},
  organization={IEEE}
}

@inproceedings{singh2021textocr,
  title={Textocr: Towards large-scale end-to-end reasoning for arbitrary-shaped scene text},
  author={Singh, Amanpreet and Pang, Guan and Toh, Mandy and Huang, Jing and Galuba, Wojciech and Hassner, Tal},
  booktitle={Proceedings of the IEEE/CVF conference on computer vision and pattern recognition},
  pages={8802--8812},
  year={2021}
}

@inproceedings{krylov2021open,
  title={Open images v5 text annotation and yet another mask text spotter},
  author={Krylov, Ilya and Nosov, Sergei and Sovrasov, Vladislav},
  booktitle={Asian Conference on Machine Learning},
  pages={379--389},
  year={2021},
  organization={PMLR}
}

@inproceedings{mishra2012scene,
  title={Scene text recognition using higher order language priors},
  author={Mishra, Anand and Alahari, Karteek and Jawahar, CV},
  booktitle={BMVC-British machine vision conference},
  year={2012},
  organization={BMVA}
}

@article{risnumawan2014robust,
  title={A robust arbitrary text detection system for natural scene images},
  author={Risnumawan, Anhar and Shivakumara, Palaiahankote and Chan, Chee Seng and Tan, Chew Lim},
  journal={Expert Systems with Applications},
  volume={41},
  number={18},
  pages={8027--8048},
  year={2014},
  publisher={Elsevier}
}

@inproceedings{wang2011end,
  title={End-to-end scene text recognition},
  author={Wang, Kai and Babenko, Boris and Belongie, Serge},
  booktitle={2011 International conference on computer vision},
  pages={1457--1464},
  year={2011},
  organization={IEEE}
}

@inproceedings{phan2013recognizing,
  title={Recognizing text with perspective distortion in natural scenes},
  author={Phan, Trung Quy and Shivakumara, Palaiahnakote and Tian, Shangxuan and Tan, Chew Lim},
  booktitle={Proceedings of the IEEE international conference on computer vision},
  pages={569--576},
  year={2013}
}

@inproceedings{karatzas2013icdar,
  title={ICDAR 2013 robust reading competition},
  author={Karatzas, Dimosthenis and Shafait, Faisal and Uchida, Seiichi and Iwamura, Masakazu and i Bigorda, Lluis Gomez and Mestre, Sergi Robles and Mas, Joan and Mota, David Fernandez and Almazan, Jon Almazan and De Las Heras, Lluis Pere},
  booktitle={2013 12th international conference on document analysis and recognition},
  pages={1484--1493},
  year={2013},
  organization={IEEE}
}

@inproceedings{karatzas2015icdar,
  title={ICDAR 2015 competition on robust reading},
  author={Karatzas, Dimosthenis and Gomez-Bigorda, Lluis and Nicolaou, Anguelos and Ghosh, Suman and Bagdanov, Andrew and Iwamura, Masakazu and Matas, Jiri and Neumann, Lukas and Chandrasekhar, Vijay Ramaseshan and Lu, Shijian and others},
  booktitle={2015 13th international conference on document analysis and recognition (ICDAR)},
  pages={1156--1160},
  year={2015},
  organization={IEEE}
}

@article{lan2012optimal,
  title={An optimal method for stochastic composite optimization},
  author={Lan, Guanghui},
  journal={Mathematical Programming},
  volume={133},
  number={1-2},
  pages={365--397},
  year={2012},
  publisher={Springer}
}

@inproceedings{smith2019super,
  title={Super-convergence: Very fast training of neural networks using large learning rates},
  author={Smith, Leslie N and Topin, Nicholay},
  booktitle={Artificial intelligence and machine learning for multi-domain operations applications},
  volume={11006},
  pages={369--386},
  year={2019},
  organization={SPIE}
}

@inproceedings{cubuk2020randaugment,
  title={Randaugment: Practical automated data augmentation with a reduced search space},
  author={Cubuk, Ekin D and Zoph, Barret and Shlens, Jonathon and Le, Quoc V},
  booktitle={Proceedings of the IEEE/CVF conference on computer vision and pattern recognition workshops},
  pages={702--703},
  year={2020}
}

@article{yang2019xlnet,
  title={Xlnet: Generalized autoregressive pretraining for language understanding},
  author={Yang, Zhilin and Dai, Zihang and Yang, Yiming and Carbonell, Jaime and Salakhutdinov, Russ R and Le, Quoc V},
  journal={Advances in neural information processing systems},
  volume={32},
  year={2019}
}

@ARTICLE{9623473,
  author={Li, Ming and Fu, Bin and Zhang, Zhengfu and Qiao, Yu},
  journal={IEEE Transactions on Multimedia}, 
  title={Character-Aware Sampling and Rectification for Scene Text Recognition}, 
  year={2023},
  volume={25},
  number={},
  pages={649-661},
  doi={10.1109/TMM.2021.3129651}}

@ARTICLE{9798797,
  author={Chen, Zhuo and Yin, Fei and Yang, Qing and Liu, Cheng-Lin},
  journal={IEEE Transactions on Multimedia}, 
  title={Cross-Lingual Text Image Recognition via Multi-Hierarchy Cross-Modal Mimic}, 
  year={2023},
  volume={25},
  number={},
  pages={4830-4841},
  doi={10.1109/TMM.2022.3183386}}

@ARTICLE{10078345,
  author={Li, Zhe and Wang, Xinyu and Liu, Yuliang and Jin, Lianwen and Huang, Yichao and Ding, Kai},
  journal={IEEE Transactions on Multimedia}, 
  title={Improving Handwritten Mathematical Expression Recognition via Similar Symbol Distinguishing}, 
  year={2024},
  volume={26},
  number={},
  pages={90-102},
  doi={10.1109/TMM.2023.3260648}}

@article{yang2024class,
  title={Class-Aware Mask-guided feature refinement for scene text recognition},
  author={Yang, Mingkun and Yang, Biao and Liao, Minghui and Zhu, Yingying and Bai, Xiang},
  journal={Pattern Recognition},
  volume={149},
  pages={110244},
  year={2024},
  publisher={Elsevier}
}

@inproceedings{xu2024ote,
  title={OTE: Exploring Accurate Scene Text Recognition Using One Token},
  author={Xu, Jianjun and Wang, Yuxin and Xie, Hongtao and Zhang, Yongdong},
  booktitle={Proceedings of the IEEE/CVF Conference on Computer Vision and Pattern Recognition},
  pages={28327--28336},
  year={2024}
}



%

\begin{IEEEbiography}
[{\includegraphics[width=1in,height=1.25in,clip,keepaspectratio]{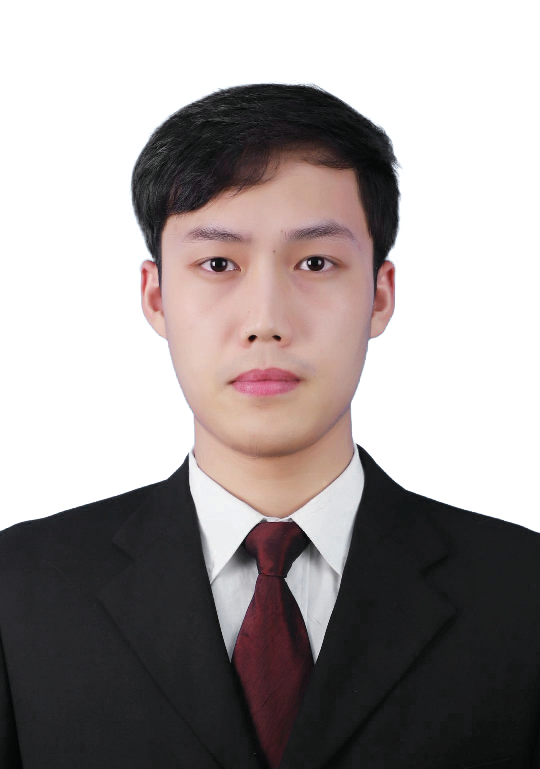}}]{Honghui Chen} received the B.E.degree in Electronic Information Engineering from Fuzhou University, Fuzhou, China, in 2020. He is currently working toward the Ph.D. degree in Information and Communication Engineering in the College of Physics and Information Engineering, Fuzhou University, Fuzhou, China. His current research interests include deep learning, and computer vision.
\end{IEEEbiography}

\begin{IEEEbiography}
[{\includegraphics[width=1in,height=1.25in,clip,keepaspectratio]{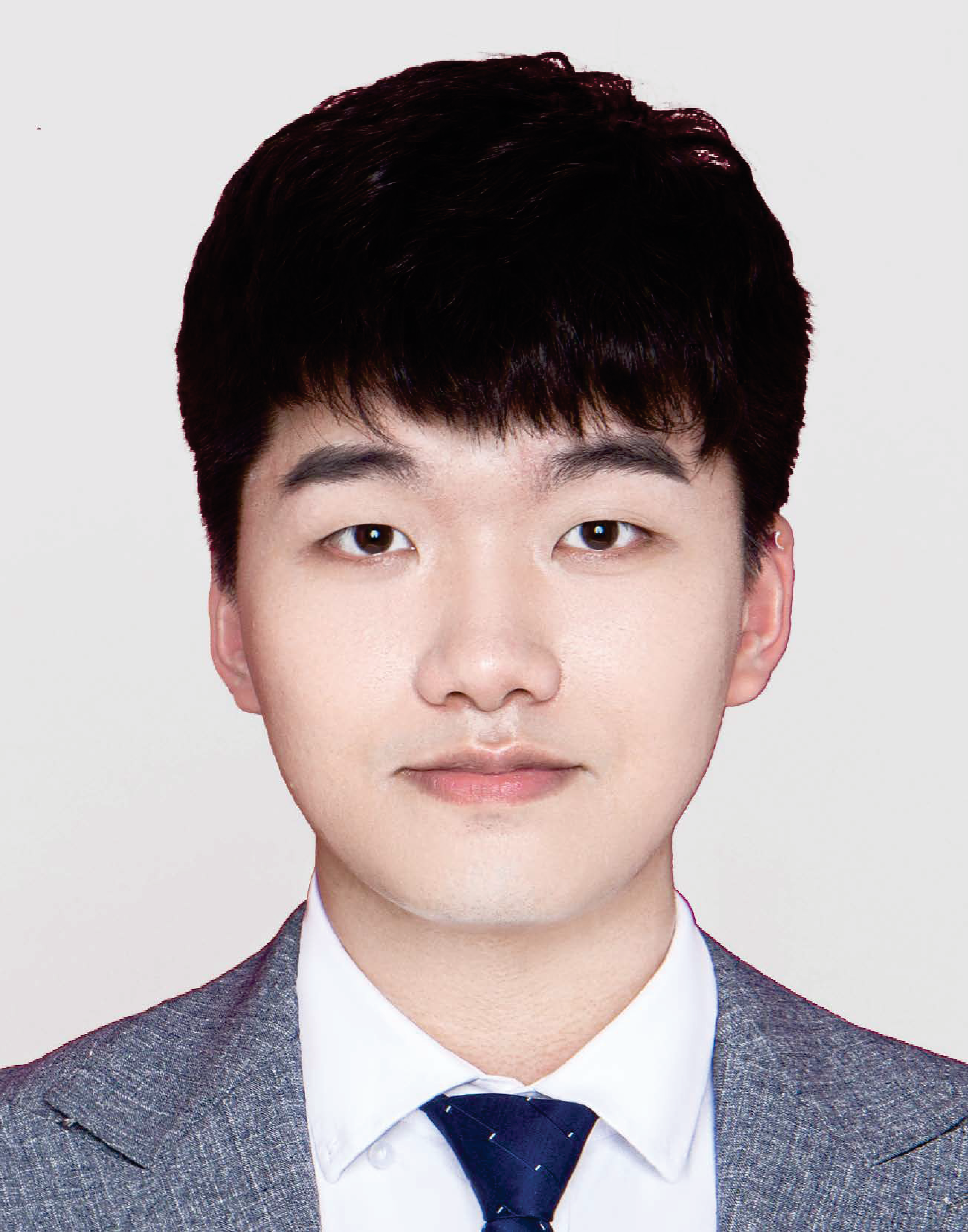}}]{Yuhang Qiu} is currently a Ph.D. student in faculty of engineering, Monash University, Australia. From June 2020 to June 2021, he was a Research Assistant in electronic and information engineering with Suzhou Institute of Biomedical Engineering and Technology, Chinese Academy of Sciences. He studied as a visiting student for half a year in National Chiao Tung University, Taiwan, in 2019. His primary research interests include deep learning, big data analysis, smart coking and biometrics.
\end{IEEEbiography}

\begin{IEEEbiography}
[{\includegraphics[width=1in,height=1.25in,clip,keepaspectratio]{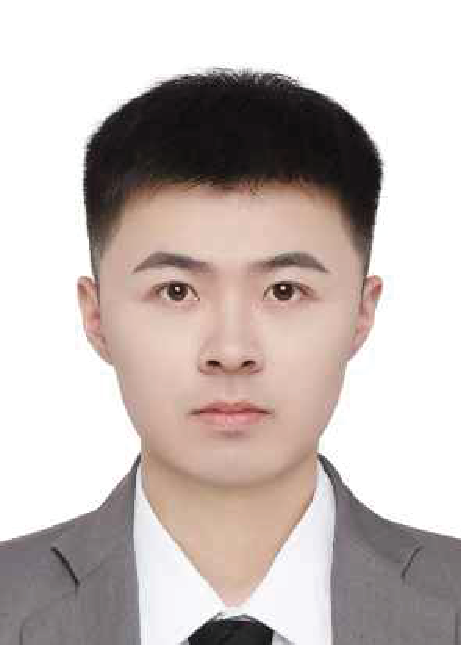}}]{Jiabao Wang} received his B.S. degree from the College of Physics and Information Engineering, Fuzhou University, in 2021. Currently, he is pursuing his master degree at the College of Physics and Information Engineering, Fuzhou University, Fuzhou, China. His research interests include computer vision and image processing.
\end{IEEEbiography}

\begin{IEEEbiography}
[{\includegraphics[width=1in,height=1.25in,clip,keepaspectratio]{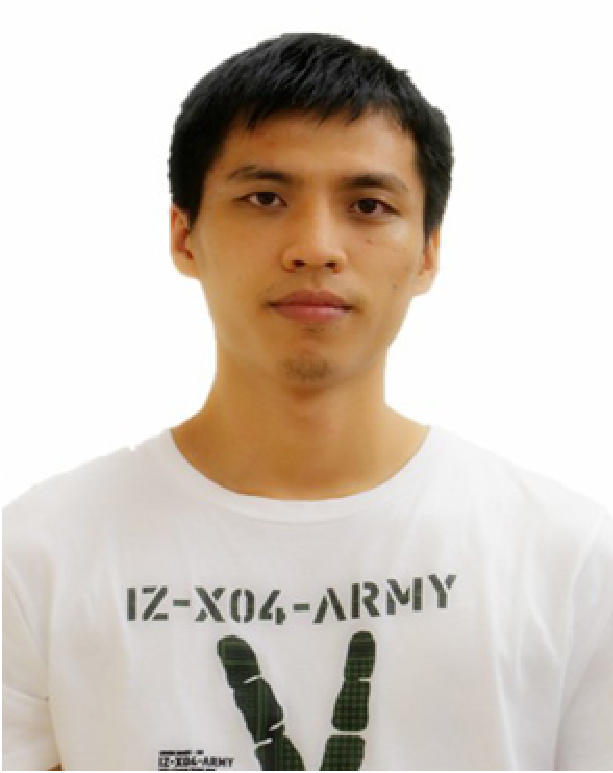}}]{Pingping Chen} (M'15--SM{'}22) received the Ph.D. degree in electronic engineering, Xiamen University, China, in 2013. From May 2012 to September 2012, he was a Research Assistant in electronic and information engineering with The Hong Kong Polytechnic University, Hong Kong. From January 2013 to January 2015, he was a Postdoctoral Fellow at the Institute of Network Coding, Chinese University of Hong Kong, Hong Kong. He is currently a Professor in Fuzhou University, China. His primary research interests include machine learning and image processing.
\end{IEEEbiography}

\begin{IEEEbiography}
[{\includegraphics[width=1in,height=1.25in,clip,keepaspectratio]{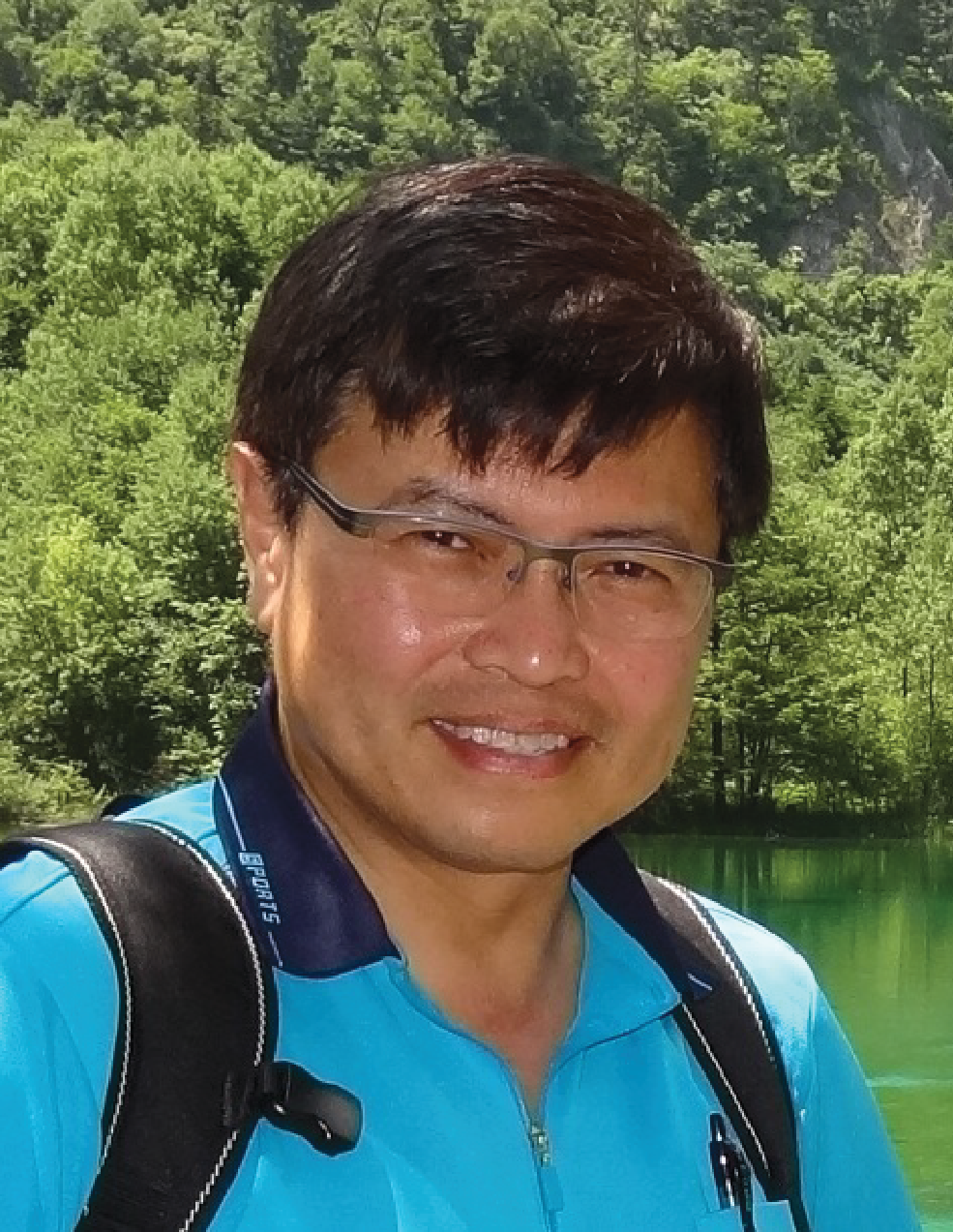}}]{Nam Ling} (S{'}88–M{'}9–SM{'}99–F{'}08–LF{'}22) received the B.Eng. degree (Electrical Engineering) from the National University of Singapore and the M.S. and Ph.D. degrees (Computer Engineering) from the University of Louisiana, Lafayette, U.S.A. He is currently the Associate Dean (Research) for the School of Engineering and the Wilmot J. Nicholson Family Chair Professor of Santa Clara University (U.S.A). From 2010-2023, he was the Chair of the Department of Computer Science \& Engineering. From 2010 to 2020, he was the Sanfilippo Family Chair Professor of Santa Clara University. From 2002 to 2010, he was also an Associate Dean for its School of Engineering. He is/was also a Chair/Distinguished/Guest/Consulting Professor for seven universities internationally. He has more than 300 publications in video/image coding and systolic arrays. He also has seven adopted standards contributions and has been granted with more than 20 U.S./European/PCT patents. He is an IEEE Fellow due to his contributions to video coding algorithms and architectures. He is also an IET Fellow. He was named IEEE Distinguished Lecturer twice and was also an APSIPA Distinguished Lecturer. He received the IEEE ICCE Best Paper Award (First Place) and the IEEE Umedia Best/Excellent Paper Awards (three times). He received six awards from Santa Clara University, four at the University
level and two at the School/College level. He has served as Keynote/Distinguished Speaker for IEEE APCCAS, VCVP (twice), JCPC, IEEE ICAST, IEEE ICIEA (twice), IET FC \& U-Media, IEEE U-Media, ICNLP/SSPS/CVPS, Workshop at XUPT (twice), and ICCIT. He is/was General Chair/Co Chair/Honorary Co-Chair for IEEE Hot Chips, VCVP (twice), IEEE ICME, IEEE VCIP, IEEE U-Media (six times), and IEEE SiPS. He has also served as Technical Program Co Chair for IEEE ISCAS (twice), IEEE ICME, APSIPA ASC, IEEE APCCAS, IEEE SiPS (twice), DCV, and IEEE VCIP. He was Technical Committee Chair for IEEE CASCOM TC and IEEE TCMM, and has served as Guest Editor/Associate Editor for IEEE TCAS I, IEEE J-STSP, Springer JSPS, Springer MSSP, and other journals. He was the Chair of the APSIPA US Chapter and organized a panel. He has delivered more than 120 invited colloquia worldwide and has served as Visiting Professor/Consultant/Scientist for many institutions/companies.
\end{IEEEbiography}







\end{document}